\documentclass[11pt,draftcls,onecolumn]{IEEEtran} 

\pdfoutput=1
\usepackage{amsmath,amssymb}
\usepackage{stmaryrd}
\usepackage{enumerate}
\usepackage{array}
\usepackage{bbm}
\usepackage{eqparbox}
\usepackage{frame}
\usepackage{float}
\usepackage{algorithm}
\usepackage{color}
\usepackage[tight,footnotesize]{subfigure}
\usepackage{url}
\usepackage{epsfig}
 \usepackage{multirow}

\newcommand{\Section}[1]{  \vspace{-0.15in} \section{#1}  \vspace{-0.1in} } 
\newcommand{\Subsection}[1]{ \vspace{-0.1in} \subsection{#1}  \vspace{-0.075in} }

\begin{document}

\title{PaFiMoCS: Particle Filtered Modified-CS and Applications in Visual Tracking across Illumination Change}

\author{Rituparna Sarkar, Samarjit Das  and Namrata Vaswani
\thanks{R. Sarkar and N. Vaswani are with the Department of Electrical and Computer Engineering, Iowa State University, Ames, IA, 50011. 
S. Das is with the Computer Science dept at Carnegie Mellon University.  
E-mail: \{rsarkar,namrata\}@iastate.edu, samarjit@cs.cmu.edu. Part of this work appeared in \cite{pafimocs_asilomar}. This work was partly funded by NSF grant IIS-1117509.
}
}

\maketitle

\newcommand{\pss}{p^{**,i}}

\newcommand{\vect}[2]{\left[\begin{array}{cccccc}
     #1 \\
     #2
   \end{array}
  \right]
  }

\newcommand{\matr}[2]{ \left[\begin{array}{cc}
     #1 \\
     #2
   \end{array}
  \right]
  }
\newcommand{\gdot}{\dot{g}}
\newcommand{\Cdot}{\dot{C}}
\newcommand{\re}{\mathbb{R}}
\newcommand{\n}{{\cal N}}  
\newcommand{\N}{{\overrightarrow{\bf N}}}  
\newcommand{\chat}{\tilde{C}_n}
\newcommand{\cmin}{C^*_{min}}
\newcommand{\twi}{\tilde{w}_n^{(i)}}
\newcommand{\twj}{\tilde{w}_n^{(j)}}
\newcommand{\wi}{{w}_n^{(i)}}
\newcommand{\twio}{\tilde{w}_{n-1}^{(i)}}

\newcommand{\tWi}{\tilde{W}_n^{(m)}}
\newcommand{\tWj}{\tilde{W}_n^{(k)}}
\newcommand{\Wi}{{W}_n^{(m)}}
\newcommand{\tWio}{\tilde{W}_{n-1}^{(m)}}

\newcommand{\ds}{\displaystyle}

\newcommand{\SAR}{S$\!$A$\!$R }
\newcommand{\MAR}{MAR}
\newcommand{\MMRF}{MMRF}
\newcommand{\AR}{A$\!$R }
\newcommand{\GMRF}{G$\!$M$\!$R$\!$F }
\newcommand{\DTM}{D$\!$T$\!$M }
\newcommand{\MSE}{M$\!$S$\!$E }
\newcommand{\RCS}{R$\!$C$\!$S }
\newcommand{\uomega}{\underline{\omega}}
\newcommand{\lu}{\mu}
\newcommand{\g}{g}
\newcommand{\s}{{\bf s}}
\newcommand{\bft}{{\bf t}}
\newcommand{\refmap}{{\cal R}}
\newcommand{\totrefl}{{\cal E}}
\newcommand{\beq}{\begin{equation}}
\newcommand{\eeq}{\end{equation}}
\newcommand{\bdm}{\begin{displaymath}}
\newcommand{\edm}{\end{displaymath}}
\newcommand{\hatz}{\hat{z}}
\newcommand{\hatu}{\hat{u}}
\newcommand{\tilz}{\tilde{z}}
\newcommand{\tilu}{\tilde{u}}
\newcommand{\hhatz}{\hat{\hat{z}}}
\newcommand{\hhatu}{\hat{\hat{u}}}
\newcommand{\tilc}{\tilde{C}}
\newcommand{\hatc}{\hat{C}}
\newcommand{\tim}{n}

\newcommand{\ssp}{\renewcommand{\baselinestretch}{1.0}}
\newcommand{\defd}{\mbox{$\stackrel{\mbox{$\triangle$}}{=}$}}
\newcommand{\goes}{\rightarrow}
\newcommand{\tends}{\rightarrow}
\newcommand{\defn}{\triangleq} 
\newcommand{\se}{&=&}
\newcommand{\sdefn}{& \defn  &}
\newcommand{\sle}{& \le &}
\newcommand{\sge}{& \ge &}
\newcommand{\plusminus}{\stackrel{+}{-}}
\newcommand{\Ey}{E_{Y_{1:t}}}
\newcommand{\ey}{E_{Y_{1:t}}}

\newcommand{\equivto}{\mbox{~~~which is equivalent to~~~}}
\newcommand{\nonzero}{i:\pi^n(x^{(i)})>0}
\newcommand{\nonzeroc}{i:c(x^{(i)})>0}

\newcommand{\supn}{\sup_{\phi:||\phi||_\infty \le 1}}
\newtheorem{definition}{Definition}
\newtheorem{remark}{Remark}
\newtheorem{ass}{Assumption}
\newtheorem{fact}{Fact}
\newcommand{\eps}{\epsilon}
\newcommand{\bd}{\begin{definition}}
\newcommand{\ed}{\end{definition}}
\newcommand{\udq}{\underline{D_Q}}
\newcommand{\td}{\tilde{D}}
\newcommand{\epsinv}{\epsilon_{inv}}
\newcommand{\al}{\mathcal{A}}

\newcommand{\bfx} {\bf X}
\newcommand{\bfy} {\bf Y}
\newcommand{\bfz} {\bf Z}
\newcommand{\ddas}{\mbox{${d_1}^2({\bf X})$}}
\newcommand{\ddbs}{\mbox{${d_2}^2({\bfx})$}}
\newcommand{\dda}{\mbox{$d_1(\bfx)$}}
\newcommand{\ddb}{\mbox{$d_2(\bfx)$}}
\newcommand{\xinc}{{\bfx} \in \mbox{$C_1$}}
\newcommand{\eqa}{\stackrel{(a)}{=}}
\newcommand{\eqb}{\stackrel{(b)}{=}}
\newcommand{\eqe}{\stackrel{(e)}{=}}
\newcommand{\leqc}{\stackrel{(c)}{\le}}
\newcommand{\leqd}{\stackrel{(d)}{\le}}

\newcommand{\leqa}{\stackrel{(a)}{\le}}
\newcommand{\leqb}{\stackrel{(b)}{\le}}
\newcommand{\leqe}{\stackrel{(e)}{\le}}
\newcommand{\leqf}{\stackrel{(f)}{\le}}
\newcommand{\leqg}{\stackrel{(g)}{\le}}
\newcommand{\leqh}{\stackrel{(h)}{\le}}
\newcommand{\leqi}{\stackrel{(i)}{\le}}
\newcommand{\leqj}{\stackrel{(j)}{\le}}

\newcommand{\w}{{W^{LDA}}}
\newcommand{\halpha}{\hat{\alpha}}
\newcommand{\hsigma}{\hat{\sigma}}
\newcommand{\slmax}{\sqrt{\lambda_{max}}}
\newcommand{\slmin}{\sqrt{\lambda_{min}}}
\newcommand{\lmax}{\lambda_{max}}
\newcommand{\lmin}{\lambda_{min}}

\newcommand{\da} {\frac{\alpha}{\sigma}}
\newcommand{\chka} {\frac{\check{\alpha}}{\check{\sigma}}}
\newcommand{\sumo}{\sum _{\underline{\omega} \in \Omega}}
\newcommand{\distance}{d\{(\hatz _x, \hatz _y),(\tilz _x, \tilz _y)\}}
\newcommand{\col}{{\rm col}}
\newcommand{\rcs}{\sigma_0}
\newcommand{\CalR}{{\cal R}}
\newcommand{\df}{{\delta p}}
\newcommand{\dq}{{\delta q}}
\newcommand{\dZ}{{\delta Z}}
\newcommand{\pprime}{{\prime\prime}}

\newcommand{\vn}{N}

\newcommand{\bv}{\begin{vugraph}}
\newcommand{\ev}{\end{vugraph}}
\newcommand{\bi}{\begin{itemize}}
\newcommand{\ei}{\end{itemize}}
\newcommand{\ben}{\begin{enumerate}}
\newcommand{\een}{\end{enumerate}}
\newcommand{\be}{\protect\[}
\newcommand{\ee}{\protect\]}
\newcommand{\bean}{\begin{eqnarray*} }
\newcommand{\eean}{\end{eqnarray*} }
\newcommand{\bea}{\begin{eqnarray} }
\newcommand{\eea}{\end{eqnarray} }
\newcommand{\nn}{\nonumber}
\newcommand{\ba}{\begin{array} }
\newcommand{\ea}{\end{array} }
\newcommand{\ep}{\mbox{\boldmath $\epsilon$}}
\newcommand{\epp}{\mbox{\boldmath $\epsilon '$}}
\newcommand{\Lep}{\mbox{\LARGE $\epsilon_2$}}
\newcommand{\und}{\underline}
\newcommand{\pdif}[2]{\frac{\partial #1}{\partial #2}}
\newcommand{\odif}[2]{\frac{d #1}{d #2}}
\newcommand{\dt}[1]{\pdif{#1}{t}}
\newcommand{\urho}{\underline{\rho}}

\newcommand{\spc}{{\cal S}}
\newcommand{\tspc}{{\cal TS}}

\newcommand{\uq}{\underline{q}^*}
\newcommand{\uv}{\underline{v}}
\newcommand{\us}{\underline{s}^*}
\newcommand{\uc}{\underline{c}}
\newcommand{\utheta}{\underline{\theta}^*}
\newcommand{\ualpha}{\underline{\alpha^*}}

\newcommand{\uxy}{\underline{x}^*}
\newcommand{\uxyj}{[x^{*}_j,y^{*}_j]}
\newcommand{\arcl}[1]{arclen(#1)}
\newcommand{\one}{{\mathbf{1}}}

\newcommand{\uxyjt}{\uxy_{j,t}}
\newcommand{\E}{\mathbb{E}}

\newcommand{\rhomat}{\left[\begin{array}{c}
                        \rho_3 \ \rho_4 \\
                        \rho_5 \ \rho_6
                        \end{array}
                   \right]}
\newcommand{\deltat}{\tau} 
\newcommand{\deltatt}{\Delta t_1}
\newcommand{\ceil}[1]{\ulcorner #1 \urcorner}

\newcommand{\xxi}{x^{(i)}}
\newcommand{\txi}{\tilde{x}^{(i)}}
\newcommand{\txj}{\tilde{x}^{(j)}}

\newcommand{\mi}[1]{{#1}^{(m,i)}}


\begin{abstract}
We study the problem of tracking (causally estimating) a time sequence of sparse spatial signals with changing sparsity patterns, as well as other unknown states, from a sequence of nonlinear observations corrupted by (possibly) non-Gaussian noise. In many applications, particularly those in visual tracking, the unknown state can be split into a small dimensional part, e.g. global motion, and a spatial signal, e.g. illumination or shape deformation. The spatial signal is often well modeled as being sparse in some domain. For a long sequence, its sparsity pattern can change over time, although the changes are usually slow. 
To address the above problem, we propose a novel solution approach called Particle Filtered Modified-CS (PaFiMoCS). The key idea of PaFiMoCS is to importance sample for the small dimensional state vector, while replacing importance sampling by slow sparsity pattern change constrained posterior mode tracking for recovering the sparse spatial signal. We show that the problem of tracking moving objects across spatially varying illumination change is an example of the above problem and explain how to design PaFiMoCS for it. Experiments on both simulated data as well as on real videos with significant illumination changes demonstrate the superiority of the proposed algorithm as compared with existing particle filter based tracking algorithms.
\end{abstract}


\begin{figure*}[!t]
\centerline{
\subfigure[full size image sequence]{
\includegraphics[height=65mm,width=70mm]{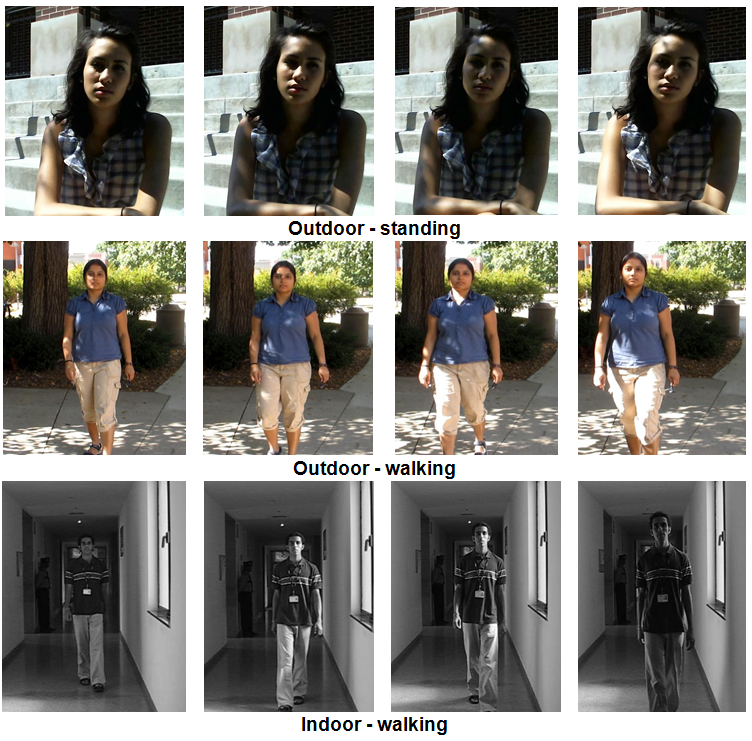}
}
\subfigure[zoom in on the faces]{
\includegraphics[height=65mm,width=70mm]{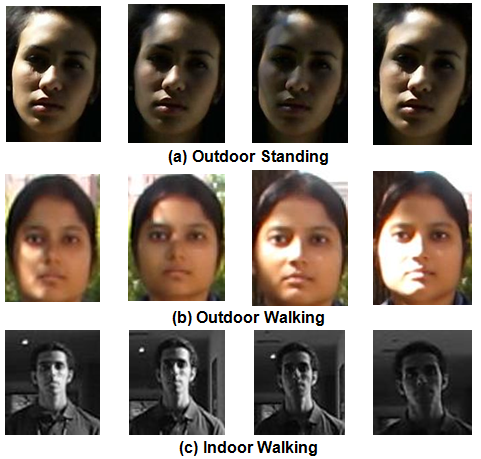}
}
}
\vspace{-0.15in}
\caption{{\small Three examples of videos with significant spatial and temporal illumination variations. 
In the first sequence (outdoor-standing), the person is standing under a large tree on a very windy day.  In the second one (outdoor-walking), the person is walking under a large tree, on a windy day. The third sequence (indoor-walking) consists of a person walking in a corridor past a window.
}} 
\vspace{-0.25in}
\label{sample_illu}
\end{figure*}

\vspace{-0.2in}

\Section{Introduction}

{\em We request the reviewers to review the single column version (draft version) since the equations are formatted for that one.}

In this work, we study the problem of tracking (causally estimating) a time sequence of sparse spatial signals with slowly changing sparsity patterns, as well as other unknown states, from a sequence of nonlinear observations corrupted by (possibly) non-Gaussian noise.  In many practical applications, particularly those in visual tracking, the unknown state can be split into a small dimensional part, e.g. global motion and a spatial signal (large dimensional part), e.g. illumination or shape deformation. The spatial signal is often well modeled as being sparse in some domain. For a long sequence, its sparsity pattern (the support set of the sparsity basis coefficients' vector) can change over time, although the changes are usually slow. 

A key example of the above problem occurs in tracking moving objects across spatially varying illumination changes, e.g. persons walking under a tree (different lighting falling on different parts of the face due to the leaves blocking or not blocking the sunlight and this pattern changes with time as the leaves move); or indoor sequences with variable lighting in different parts of the room, either due to the placement of light sources, or due to sunlight coming in through the windows that illuminates certain parts of the room better than others. For some examples, see Fig \ref{sample_illu}.
In all of these cases, one needs to explicitly track the motion (small dimensional part) as well as the illumination ``image" (illumination at each pixel in the image). As we explain in Sec \ref{contrib}, our work is the {\em first} to demonstrate that the illumination image sequence for many real videos can indeed be modeled as being sparse with slowly changing sparsity patterns.
Another example application is tracking the boundary contours of moving and deforming objects over time. Here again motion constitutes the small dimensional part, whereas deformation is the large dimensional spatial signal that can often be modeled as being Fourier sparse.
%
%

\Subsection{Related Work} 
In recent years, starting with the seminal papers of Candes, Romberg, Tao and of Donoho \cite{candes,donoho} there has been a large amount of work on sparse signal recovery or what is now more commonly known as compressive sensing (CS).  The problem of recursively recovering a time sequence of sparse signals, with slowly changing sparsity patterns and signal values, from linear measurements was introduced in \cite{kfcsicip, just_lscs} where a solution called Kalman filtered CS (KF-CS) was proposed, and a simplification of KF-CS called Least Squares CS (LS-CS) was rigorously analyzed. Later work on the topic includes modified-CS \cite{isitmodcs}, weighted $\ell_1$ \cite{hassibi}, modified block CS \cite{modblockcs_stojnic} and regularized modified-CS \cite{regmodbpdn}.
 Recent Bayesian approaches to sequential sparse estimation with time-varying supports include \cite{sparse_dynamic_sys,schniter_track,zhang_rao,romberg_ciss11,unscented_cs}. However, all of these works propose solution approaches for problems with linear measurement models.

For tracking problems that need to causally estimate a time sequence of hidden states, $X_t$, from nonlinear and possibly non-Gaussian measurements, $Y_t$,  the most common and efficient solution is to use a particle filter (PF). A PF can be used if the sequence $\{X_t,Y_t\}$ satisfies the hidden Markov model assumption. The PF uses sequential importance sampling \cite{doucet} along with a resampling step \cite{gordon} to obtain a sequential Monte Carlo estimate of the posterior distribution, $\pi_{t|t}(x_t) := f_{X_t|Y_{1:t}}(x_t|y_{1:t})$, of the state $X_t$ conditioned on all observations up to the current time, $Y_{1:t}$.

In the problem that we study, part of the state vector is a discrete spatial signal and hence very high dimensional. As a result, in this case, the original PF \cite{gordon} will require too many particles for accurate tracking and hence becomes impractical to use. As explained in \cite{das_tip_11}, the same is essentially true for most existing PF algorithms. Some of the efficient PFs such as PF-Doucet\cite{doucet}, Gaussian PF \cite{gpf}, Gaussian sum filters or Gaussian sum PF \cite{gspf} also cannot be used for the following reason. The first two implicitly assume that the posterior conditioned on the previous state, $f_{X_t|Y_{t},X_{t-1}}(x_t|y_{t},x_{t-1})$, is unimodal or is at least unimodal most of the time. The second two assume a linear, or at least, a unimodal, observation model. In our problem, the observation model is nonlinear and is such that it often results in a multimodal observation likelihood, e.g., as explained in \cite{das_tip_11}, this happens due to background clutter for the illumination-motion tracking problem. If, in addition, the state transition prior of the small dimensional state, e.g., the motion states, is broad, which is often the case, it will result in $f_{X_t|Y_{t},X_{t-1}}(x_t|y_{t},x_{t-1})$ being multimodal. This fact is explained in \cite{das_tip_11} for the illumination-motion problem. Moreover, if the nonlinearity is such that the state to observation mapping is not differentiable, then one cannot even find the mode of $f_{X_t|Y_{t},X_{t-1}}(x_t|y_{t},x_{t-1})$ and hence cannot even implement PF-Doucet. This is again true for the illumination-motion problem. Frequently multimodal observation likelihoods and the above non-differentiability also mean that the extended Kalman filter, the unscented Kalman filter, the interacting multiple mode filter or Gaussian mixture filters cannot be used \cite{mcip}.

Rao-Blackwellized PF (RB-PF)  \cite{gust_nord, chen_liu} and PF with posterior mode tracking (PF-MT) algorithm \cite{pfmtpap} are two possible solutions for large dimensional tracking problems.
RB-PF requires that conditioned on the small dimensional state vector, the state space model be linear and Gaussian. PF-MT relaxes this and only requires that, conditioned on the previous state and a small dimensional state vector, the posterior of the rest of the state vector (large dimensional part) be unimodal most of the time.  However, neither RB-PF nor PF-MT exploits the sparsity of the spatial signal to be tracked. The same is true for more recent works on large dimensional tracking \cite{chang_TIP2008,mihaylova_ICASSP2011,mihaylova_TITS2012,partas_2010},  as well as for works that introduce efficient resampling strategies \cite{Djuric_ICASSP2011,zhang_ICASSP2012}.


Other recent works in visual tracking that utilize sparsity in some fashion include \cite{sparse_kernel_TIP2010,ahuja_CVPR2012,compressive_tracking_ECCV2012,VisualTrackingl1,VisualTrackingCS,zhuang_eccv10}.
The work of \cite{sparse_kernel_TIP2010} introduces the use of sparse kernel density estimation techniques to obtain a kernel density estimate for the prior and posterior densities estimated by the PF at each time. \cite{ahuja_CVPR2012} uses sparsity for multi-task tracking by combining ideas from multi-task learning, particle filtering and $\ell_1$ minimization. The work of \cite{compressive_tracking_ECCV2012} projects the image into a lower dimensional space using a random measurement matrix and uses these in a Bayesian classifier for separating the target from the background at each time. 
The works of \cite{VisualTrackingl1, VisualTrackingCS} use a PF for target tracking, while using sparse recovery to obtain the best template out of a class of templates obtained from training data, target particles from previous time and trivial templates (identity matrix designed for occlusions and corruptions). 
The work of \cite{zhuang_eccv10} uses dynamic group sparsity for robust and fast tracking. 



\Subsection{Our Contribution}
\label{contrib}
With the exception of \cite{kfcsicip,just_lscs,isitmodcs,regmodbpdn}, none of the other works discussed above exploits the fact that, in most large dimensional tracking problems, at any given time, the large dimensional state vector is usually a spatial signal that is sparse in some dictionary/basis, and over time, its sparsity pattern can change, but the changes are usually quite slow. For example, as shown in  \cite{just_lscs,isitmodcs}, this is true for the wavelet coefficients of various dynamic MRI sequences. In this work, we show that the above is also true for the illumination image sequence.
\ben
\item  We exploit the above fact to propose a PF based tracking algorithm called Particle Filtered Modified-CS or PaFiMoCS. Unlike \cite{kfcsicip,just_lscs,isitmodcs,regmodbpdn} which only solve the linear measurements' model case, PaFiMoCS is designed for tracking problems with highly nonlinear, and possibly non-Gaussian, observation models that result in frequently multimodal observation likelihoods. Many visual tracking problems, e.g. tracking moving objects across spatially varying illumination change, fit into this category. PaFiMoCS is inspired by RB-PF and PF-MT. Its key idea is to importance sample from the state transition prior for the small dimensional state vector, while replacing importance sampling by slow sparsity pattern change constrained posterior mode tracking for recovering the sparse spatial signal. We introduce simple, but widely applicable, priors for modeling state transitions of sparse signal sequences with changing sparsity patterns that result in an efficient PaFiMoCS algorithm.

%

\item In many recent works  \cite{weiss_iccv,kalecvpr,das_tip_11}, the illumination image is often represented using the first few lowest order Legendre polynomials. However, our experiments with the dictionary of Legendre polynomials (henceforth referred to as the Legendre dictionary) are the first to demonstrate that for many video sequences with significant illumination variations, the illumination image is approximately sparse in this dictionary and, in fact, its sparsity pattern includes many of the higher order Legendre polynomials, and may not include all the lower order ones. Moreover, over time, usually the sparsity pattern changes quite slowly [see Sec \ref{supp_change_sec})].

\item We show how to design PaFiMoCS for tracking moving objects across spatially varying illumination changes. 
    Experiments on both simulated data as well as on real videos involving significant illumination changes demonstrate the superiority of the proposed algorithms.
\een

For tracking moving objects across spatially varying illumination changes, we use a template-based model taken from \cite{kalecvpr,das_tip_11} with a three-dimensional motion model, that only models x-y translation and scale. We use this because it is simple to use and to explain our key ideas.  However, we should point out that the proposed algorithm can very easily be adapted to other representations of the target e.g. feature based approaches. A similar approach can be also be developed to jointly handle appearance change due to illumination as well as other factors like 3D pose change, by using the more sophisticated models of recent work \cite{amitRC_2007,yuan_2008,kumar_2008}. Similarly, illumination can also be represented using other parameterizations such as those proposed in \cite{jacobs_jacobs, belmur_cone, hager_bell,ramamoorthi_pca}.

\Subsection{Paper Organization}
The rest of the paper is organized as follows. We give the notation and problem formulation in Sec \ref{problem}. Sec \ref{statetrans} introduces simple but widely applicable priors for modeling state transitions of sparse signal sequences with changing sparsity patterns.
We develop the PaFiMoCS algorithms in Sec \ref{pafimocs_sec}. 
The illumination-motion tracking problem and PaFiMoCS algorithms for it are described in Sec \ref{illum}. 
Experimental results, both on simulated and real video sequences, are given in Sec \ref{expts}. Conclusions and future work are discussed in Sec \ref{conclude}.

\Section{Notation, Problem Formulation and State Transition Models}
\label{problem}

\Subsection{Notation}
The probability density function (PDF) of a random variable (r.v.) $Y$ is denoted by $f_Y(y)$ while the conditional PDF of of r.v. $Y$ given another r.v. $X$ is denoted by $f_{Y|X}(y|x)$. The notations $\E[Y]$ and $\E[Y|X]$ are used to denote the expectation of $Y$ and the conditional expectation of $Y$ given $X$ respectively. The subscript $t$ denotes the discrete time index.
The notation $Z_t \stackrel{\text{i.i.d.}}{\sim} f(z)$ means that the sequence of r.v.'s $Z_1, Z_2, \dots Z_t, Z_{t+1}, \dots $ are mutually independent and identically distributed (i.i.d.) with PDF $f(z)$. If the sequence of $Z_t$'s are discrete r.v.'s then the same convention applies with PDF replaced by probability mass function (PMF).
If $Z_t$ is an $n$ length vector, the notation $(Z_t)_i \stackrel{\text{i.i.d.}}{\sim} f(z)$ means that $(Z_t)_i$'s for all $i=1,2,\dots n$ and for all $t=1,2,\dots,t,t+1,\dots$ are i.i.d. with PDF or PMF $f(z)$.

The notation $\mathcal{N}(a;\mu,\Sigma)$ denotes the value of a Gaussian PDF with mean $\mu$ and covariance matrix $\Sigma$ computed at $a$. The notation $X \sim \mathcal{N}(\mu,\Sigma)$ means that the r.v. $X$ is Gaussian distributed with mean $\mu$ and covariance matrix $\Sigma$. Similarly, the notation $\text{Unif}(a; c_1,c_2)$ denotes the value of a uniform PDF defined over $[c_1,c_2]$ computed at $a$ while $X \sim \text{Unif}(c_1,c_2)$  means that the r.v. $X$ is uniformly distributed in the interval $[c_1,c_2]$. The notation  $S \sim \text{Ber}(T,p)$ means that the set $S$ many contain any element of the set $T$ with probability $p$ (and may not contain it with probability $1-p$) independent of all the other elements of $T$.

The notation $\|{b}\|_k$ is used to denote the $\ell_k$ norm of vector ${b}$. For any set ${T}$ and vector ${b}$, $({b})_{T}$ is used to denote a sub-vector containing the elements of $b$ with indices in $ {T}$. For a matrix $A$, $ (A)_T$ denotes the sub-matrix obtained by extracting columns of $  A$ with indices in $ {T}$. We denote the complement set as $T^c$ i.e., $T^c:=\{i: i \notin T\}$. The symbols $\cup$ and $\cap$ denote set-union and set-intersection respectively and the symbol $\setminus$ denotes the set difference, i.e. $T_1 \setminus T_2:= T_1 \cap T_2^c$. For a set $T$, $|T|$ denote the cardinality of a set, but for a scalar $x$, $|x|$ denotes the magnitude of $x$.

The notation $\text{vec}(.)$ denotes the vectorization operation, it operates on a $m \times n$ matrix to give a vector of size $mn$ by cascading the rows. The Hadamard product (the $'.\ast '$ operation in MATLAB) is denoted by $\odot$. The function $\text{round}(Z)$ operates element-wise on a vector or matrix $Z$ to output a vector or matrix with integer entries closest to the corresponding elements of the vector or matrix. We use {\bf I} to denote the identity matrix and we use {\bf 1} or {\bf 0} to denote vectors with all entries as 1 or 0 respectively. $M^\top$ denotes the transpose of a matrix $M$.

\Subsection{Problem Formulation}
The goal of this work is to develop algorithms to recursively recover a time sequence of states, $X_t$, from noise-corrupted and nonlinear measurements, $Y_t$, when the state $X_t$ can be split into two parts, a large dimensional part, $L_t$, and a small dimensional part, $U_t$, with the following properties
\ben
\item $L_t$ is a discrete spatial signal, that is sparse in some known dictionary, and 
\item the sparsity pattern of $L_t$ changes slowly over time. 
\een
Mathematically, this means the following. The state $X_t$ can be split as
\bea
X_t = \left[ \begin{array}{c}
U_t \nn \\
L_t \nn
\end{array}
\right]
\eea
where $(U_t)_{n_u \times 1}$ is a small dimensional state vector and $(L_t)_{n_l \times 1}$, with $n_l \gg n_u$, is a discrete spatial signal that is sparse in some known dictionary, $\Phi$, i.e.
\bea
L_t = \Phi \Lambda_t
\eea
where $(\Lambda_t)_{n_\lambda \times 1}$ is a sparse vector with support set $T_t$, i.e.
\bea
T_t: = \text{support}(\Lambda_t) =  \{j: (\Lambda_t)_j \neq 0 \}.
\eea
The $n_l \times n_\lambda$ dictionary matrix $\Phi$ can be tall, square or fat. Often in video applications, $n_l$ is very large, and hence for computational reasons, one uses a tall dictionary matrix $\Phi$.

Notice that if $T_t$ and $(\Lambda_t)_{T_t}$ are known, then $L_t$ is known. {\em Thus, one can as well let the state vector be $X_t= [U_t^\top, T_t^\top, {(\Lambda_t)_{T_t}}^\top]^\top$ or for simplicity, just  $X_t= [U_t^\top, T_t^\top, \Lambda_t^\top]^\top$. We will use this definition of the state vector in the rest of this paper.}

The $m$ dimensional observation vector, $Y_t$, satisfies
\bea
Y_t: = h(U_t,L_t) + Z_t, \ Z_t \stackrel{\text{i.i.d.}}{\sim} f_Z(z)
\label{obsmod}
\eea
i.e. the observation noise, $Z_t$, is independent and identically distributed (i.i.d.) over time, with PDF at any time given by $f_Z(z)$. In many situations, this is Gaussian. However, often to deal with outliers, one models $Z_t$ as a mixture of two Gaussian PDF's, one which has small variance and large mixture weight and the second with large variance but small mixture weight.
For the above model, the observation likelihood, $\text{OL}(U_t,L_t)$, can be written as
\bea
\text{OL}(U_t,L_t): = f_{Y_t|U_t,L_t}(Y_t|U_t,L_t) =  f_Z(Y_t - h(U_t,L_t))
\eea
More generally, sometimes the observation model can only be specified implicitly, i.e. it can only be written in the form
\bea
g(Y_t,U_t,L_t) = Z_t, \ Z_t \stackrel{\text{i.i.d.}}{\sim} f_Z(z)
\label{obsmod_gen}
\eea
As we see in Sec \ref{illum}, this is the case for the illumination-motion tracking problem. For (\ref{obsmod_gen}), the observation likelihood, $\text{OL}(U_t,L_t)$, becomes
\bea
\text{OL}(U_t,L_t) =  f_Z(g(Y_t,U_t,L_t))
\label{ol_gen}
\eea
Notice that (\ref{obsmod}) is a special case of (\ref{obsmod_gen}) with $g(Y_t,U_t,L_t) = Y_t - h(U_t,L_t)$.


\Subsection{State Transition Models}
\label{statetrans}
In the absence of very specific model information, one can adopt the following simple state transition models. These can be used to impose slow sparsity pattern change as well as slow signal value change.

We assume the following simple support change model for $T_t$. 
\bea
T_t &=& T_{t-1} \cup A_t \setminus R_t, \ \text{where} \nn \\
A_t & \stackrel{\text{i.i.d.}}{\sim} &  \text{Ber}(T_{t-1}^c, p_a) \nn \\
R_t & \stackrel{\text{i.i.d.}}{\sim} &  \text{Ber}(T_{t-1}, p_r)
\label{sysmod1}
\eea
Here $A_t$ denotes the set added to the support at time $t$, while $R_t$ denotes the set that is removed from the support at time $t$. Each of them is i.i.d. over time. Also,  given $T_{t-1}$, $A_t$ and $R_t$ are assumed to be independent of each other. Since $A_t \subseteq T_{t-1}^c$, and $R_t \subseteq T_{t-1}$, thus $A_t$ and $T_{t-1}$ are disjoint and $|T_t| = |T_{t-1}| - |R_t| + |A_t|$.  Thus, $\E[|T_t|] = \E[|T_{t-1}|] + \E[|A_t|] - \E[|R_t|]$. Also notice that, if $S \sim \text{Ber}(T,p)$, then $\E[S | T] = |T| p$ and so $\E[S] = \E[\E[S | T]] = \E[|T|] p$. Thus,  $\E[|A_t|] = (n_\lambda - \E[|T_{t-1}|]) p_a$ and $\E[|R_t|] = \E[|T_{t-1}|] p_r$.

In most applications, it is valid to assume that the expected support size remains constant, i.e. $\E[|T_t|] = \E[|T_{t-1}|] = s$. This is ensured by setting $p_r = (n_\lambda - s) p_a / s$ so that $\E[|A_t|] = \E[|R_t|]$. Also, {\em slow support change} means that $\E[|R_t|] = \E[|A_t|]$ is small compared to $\E[|T_t|] =s$. This is ensured by picking $p_a$ to be small compared with $s/ (n_\lambda - s)$.

In the absence any of other model information, we assume the following linear Gaussian random walk model on $(\Lambda_t)_{T_t}$:
\bea
(\Lambda_t)_{T_t} &=& (\Lambda_{t-1})_{T_t} + (\nu_{l,t})_{T_t}, \ (\nu_{l,t})_{T_t} \stackrel{\text{i.i.d.}}{\sim}  \n(0,\sigma_l^2 I) \nn \\
(\Lambda_t)_{T_t^c} &=& 0
\label{sysmod2}
\eea
%
Similarly, in the absence of any other specific information, we also assume a similar model on $U_t$, i.e.
\bea
U_t &=& U_{t-1} + \nu_{u,t}, \ \nu_{u,t}   \stackrel{\text{i.i.d.}}{\sim}  \n(0,\Sigma_u)
\label{sysmod3}
\eea
If the only thing that is known is that the values of $(\Lambda_t)_{T_t}$ and $U_t$ change slowly, then the above linear Gaussian  random walk model is the most appropriate one. However, as far as the proposed algorithms are concerned, they are also applicable with minor changes for the case where $(\Lambda_t)_{T_t} = q(\Lambda_{t-1},\nu_{l,t})$ for any arbitrary but known function $q(.)$.

With the above model, the  state transition prior, $f_{X_t|X_{t-1}}(X_t^i|X_{t-1}^i)$, corresponding to the above state transition models can be written as follows. Recall that $X_t = [U_t, T_t, \Lambda_t]$.
\bea
\text{STP}(X_t^i; X_{t-1}^i) &:=& f_{X_t|X_{t-1}}(X_t^i|X_{t-1}^i)  \nn \\
                             &=& \text{STP}(T_t^i; T_{t-1}^i) \text{STP}(\Lambda_t^i;\Lambda_{t-1}^i,T_{t}^i)  \ \text{STP}(U_t^i ; U_{t-1}^i), \ \ \text{where} \\
\label{stp_T}
\text{STP}(T_t^i; T_{t-1}^i) &:= & Pr(T_t = T_t^i | T_{t-1} = T_{t-1}^i)  \nn \\
 \se Pr(A_t = (T_t^i \setminus T_{t-1}^i) |T_{t-1}= T_{t-1}^i, R_t = (T_{t-1}^i \setminus T_{t}^i) |T_{t-1}= T_{t-1}^i ) \nn \\
 \se p_a^{|T_t^i \setminus T_{t-1}^i|} (1-p_a)^{n_l - |T_{t-1}^i| - |T_t^i \setminus T_{t-1}^i|} \ p_r^{|T_{t-1}^i \setminus T_{t}^i|} (1-p_r)^{|T_{t-1}^i| - |T_{t-1}^i \setminus T_{t}^i|} \\
\label{stp_lam}
\text{STP}(\Lambda_t^i ;\Lambda_{t-1}^i, T_{t}^i) &:=& f_{\Lambda_t|\Lambda_{t-1},T_t}(\Lambda_t^i|\Lambda_{t-1}^i,T_{t}^i) \nn \\
                                                     &=&  \n((\Lambda_t^i)_{T_{t}^i} ; (\Lambda_{t-1}^i)_{T_{t}^i} , \sigma_l^2 I)  \\
\text{STP}(U_t^i ;U_{t-1}^i) &:=& f_{U_t|U_{t-1}}(U_t^i|U_{t-1}^i)   \nn \\
                              &=&  \n(U_t^i; U_{t-1}^i, \Sigma_u)
\label{stp_u}
\eea
In the above, (\ref{stp_T}) follows using the following facts: (i) $A_t = T_t \setminus T_{t-1}$ and $R_t = T_{t-1} \setminus T_{t}$; (ii) if $S \sim \text{Ber}(T,p)$, then $Pr(S=S^i|T=T^i) = p^{|S^i|} (1-p)^{|T^i|-|S^i|}$; and (iii) $A_t$ and $R_t$ are independent given $T_{t-1}$. Thus, from (\ref{sysmod1}), $Pr(A_t = A^i |T_{t-1}= T^i) = p_a^{|A^i|} (1-p_a)^{|(T^i)^c|-|A^i|}$ and similarly, $Pr(R_t = R^i |T_{t-1}= T^i) = p_r^{|R^i|} (1-p_r)^{|T^i|-|R^i|}$.
Also, (\ref{stp_lam}) and (\ref{stp_u}) follow directly from (\ref{sysmod2}) and (\ref{sysmod3}) respectively.

\Section{Particle Filtered Modified-CS}
\label{pafimocs_sec}

As explained in the introduction, most existing PF algorithms cannot be used for our problem, since we would like to deal with (a) multimodal observation likelihoods and with (b) the state consisting of a sparse spatial signal with unknown and slowly changing sparsity patterns, in addition to another small dimensional vector. Below, in Sec \ref{pfmt_review}, we provide a quick review of particle filter with mode tracker (PF-MT) and discuss its limitations. Next, in Sec \ref{pafimocs_algo2}, we explain how to address these limitations. The result is our proposed algorithm called particle filtered modified-CS (PaFiMoCS). PaFiMoCS-slow-support-change is described in Sec \ref{pafimocs_algo1}. In Sec \ref{convexity}, we give one set of the assumptions under which the cost function to be minimized by PaFiMoCS is convex. We discuss the computational cost of PaFiMoCS and the potential for its parallel implementation in Sec \ref{parallel}.
Two approaches to deal with outliers, e.g. occlusions, are described in Sec \ref{outliers}.

\begin{algorithm}
\caption{\small{PF-MT: Particle Filter with posterior Mode Tracker}}
\label{pfmt_algo}
For all $t \ge 0$ do
\ben
\item For each particle $i=1,2,\dots n_{pf}$:  Importance sample $U_t$ from its prior: $U_t^i \sim \n(0, \Sigma_u)$

\item For each particle $i=1,2,\dots n_{pf}$:  Mode track $\Lambda_t$, i.e. compute the mode of the posterior of $\Lambda_t$ conditioned on $X_{t-1}^i$ and $U_t^i$, i.e. compute $\Lambda_t^i$ as the solution of
\bea
&& \min_{\Lambda}  C(\Lambda) \ \text{where}   \  C(\Lambda): = -\log \text{OL}(U_t^i,\Phi \Lambda) + \frac{\|\Lambda - \Lambda_{t-1}^i\|_2^2}{2 \sigma_l^2}
\label{pfmt_cost}
\eea
where $\text{OL}(.)$ is defined in (\ref{ol_gen}).

\item For each particle $i=1,2,\dots n_{pf}$: Compute the weights as follows.
%
$$w_t^i \propto w_{t-1}^i \   \text{OL}(U_t^i,\Phi \Lambda) \ \n(\Lambda_t^i; \Lambda_{t-1}^i, \sigma_l^2 I)$$  


\item Resample and reset weights to $1/n_{pf}$. Increment $t$ and go to step 1.
\een
\end{algorithm}

\Subsection{A review of particle filter with mode tracker (PF-MT) and its limitations}
\label{pfmt_review}
The PF-MT algorithm \cite{pfmtpap} splits the state vector $X_t$ into $X_t = [X_{t,s}^\top, X_{t,r}^\top]^\top$ where $X_{t,s}$ denotes the coefficients of a small dimensional state vector, which can change significantly over time, while $X_{t,r}$ refers to the rest of the states which usually change much more slowly over time. PF-MT importance samples only on $X_{t,s}$, while replacing importance sampling by deterministic posterior Mode Tracking (MT) for $X_{t,r}$. Thus the importance sampling dimension is equal to only the dimension of $X_{t,s}$ which is much smaller than that of $X_t$. The importance sampling dimension is what decides the effective particle size and thus PF-MT helps to significantly reduce the number of particles needed for accurate tracking in a large dimensional problem. PF-MT implicitly assumes that (i) the posterior of $X_{t,r}$ conditioned on the previous state, $X_{t-1}$ and on $X_{t,s}$ (``conditional posterior"), i.e. $f_{X_{t,r}|X_{t-1},X_{t,s},Y_t}(X_{t,r}|X_{t-1},X_{t,s},Y_t)$, is unimodal most of the time; and that (ii) it is also narrow enough. Under these two assumptions, it can be argued that any sample from the conditional posterior is  close to its mode with high probability \cite[Theorem 2]{pfmtpap}.

PF-MT can be applied to our problem if we replace (\ref{sysmod1}) and (\ref{sysmod2}) by (\ref{sysmod2}) with $T_t = [1,2 \dots n_\lambda]$, i.e. we do not use the sparsity of $\Lambda_t$.
Then, with $X_{t,s} = U_t$ and $X_{t,r} = \Lambda_t$, the PF-MT algorithm is as given in Algorithm \ref{pfmt_algo}. Notice that the conditional posterior in this case satisfies
$$f_{\Lambda_t|X_{t-1},U_t,Y_t}(\Lambda_t|X_{t-1}^i,U_t^i,Y_t) \propto \text{OL}(U_t^i,\Phi \Lambda_t) \ \n(\Lambda_t; \Lambda_{t-1}^i, \sigma_l^2 I).$$
where $\text{OL}(.)$ is defined in (\ref{ol_gen}).  
Thus, the cost function to be minimized in the MT step is given by (\ref{pfmt_cost}).

\begin{figure*}
\centerline{
\begin{tabular}[c]{cccc}
\includegraphics[height=40mm,width=42mm]{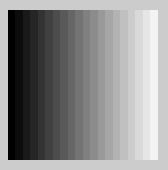}
&\includegraphics[height=40mm,width=42mm]{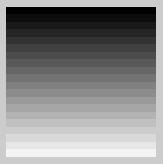}
&\includegraphics[height=40mm,width=42mm]{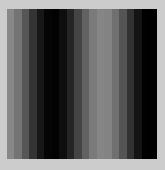}
&\includegraphics[height=40mm,width=42mm]{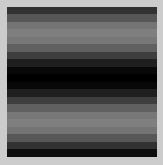}
\\
$k=2$ & $k=3$ & $k=20$ & $k=23$
\end{tabular}
}
\caption{{\small We show images of $P_k(i,j)$ defined in (\ref{def_phi}) for a few different values of $k$. Notice that higher values of $k$ correspond to higher spatial frequency of intensity change. Also, even values of $k$ correspond to x direction variation (constant along y axis) whereas odd values of $k$ corresponds to y direction variation.
}}
\vspace{-0.15in}
\label{leg_illu}
\end{figure*}

\begin{figure*}[!t]
\centerline{
\subfigure[support size]{
\label{suppsizeplot}
\begin{tabular}{c}
\includegraphics[height=55mm,width=50mm]{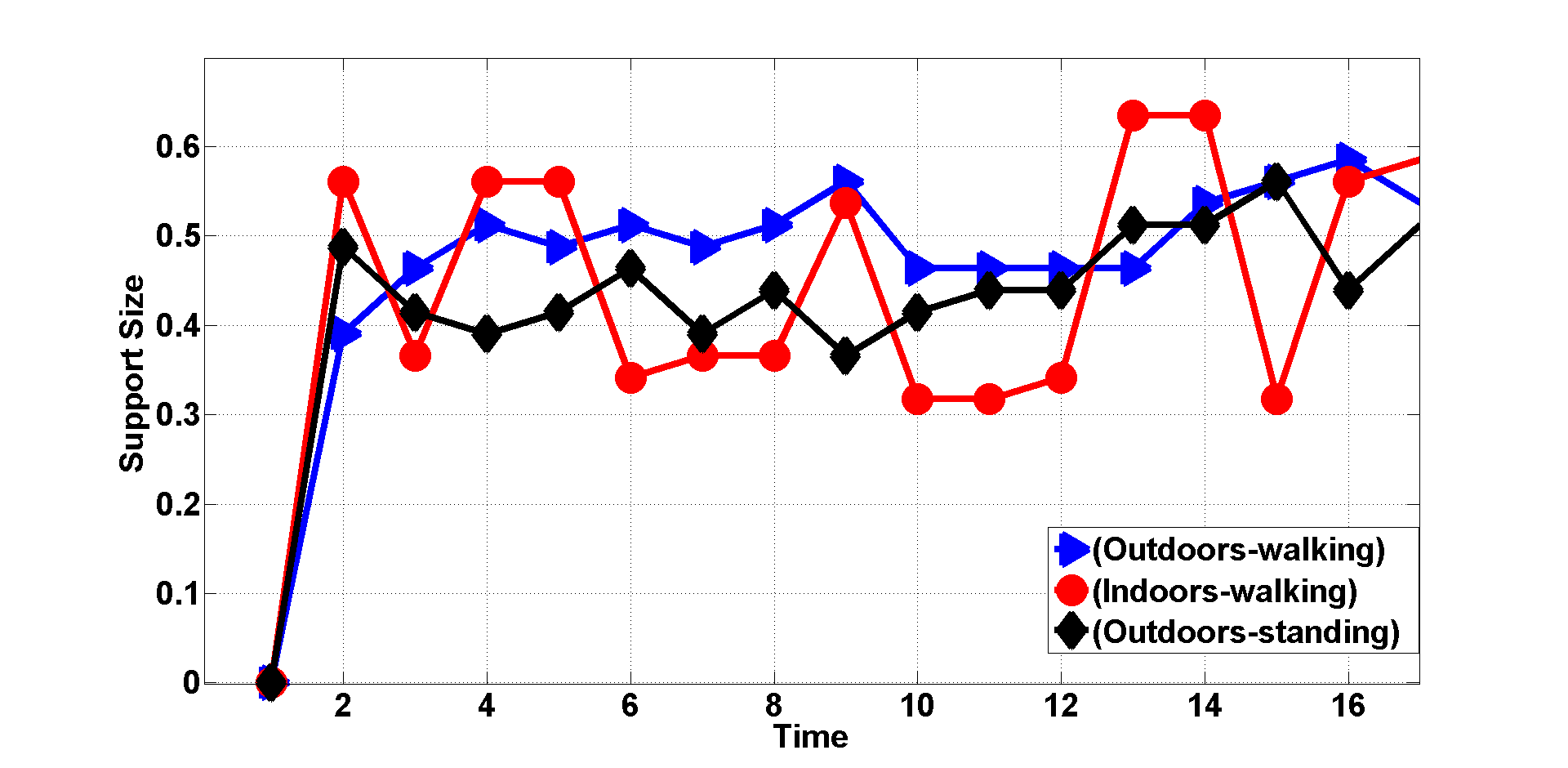}
\end{tabular}
}
\subfigure[support change sizes]{
\label{suppchangeplot}
\begin{tabular}{ccc}
\hspace{-0.4in}
\includegraphics[height=50mm,width=50mm]{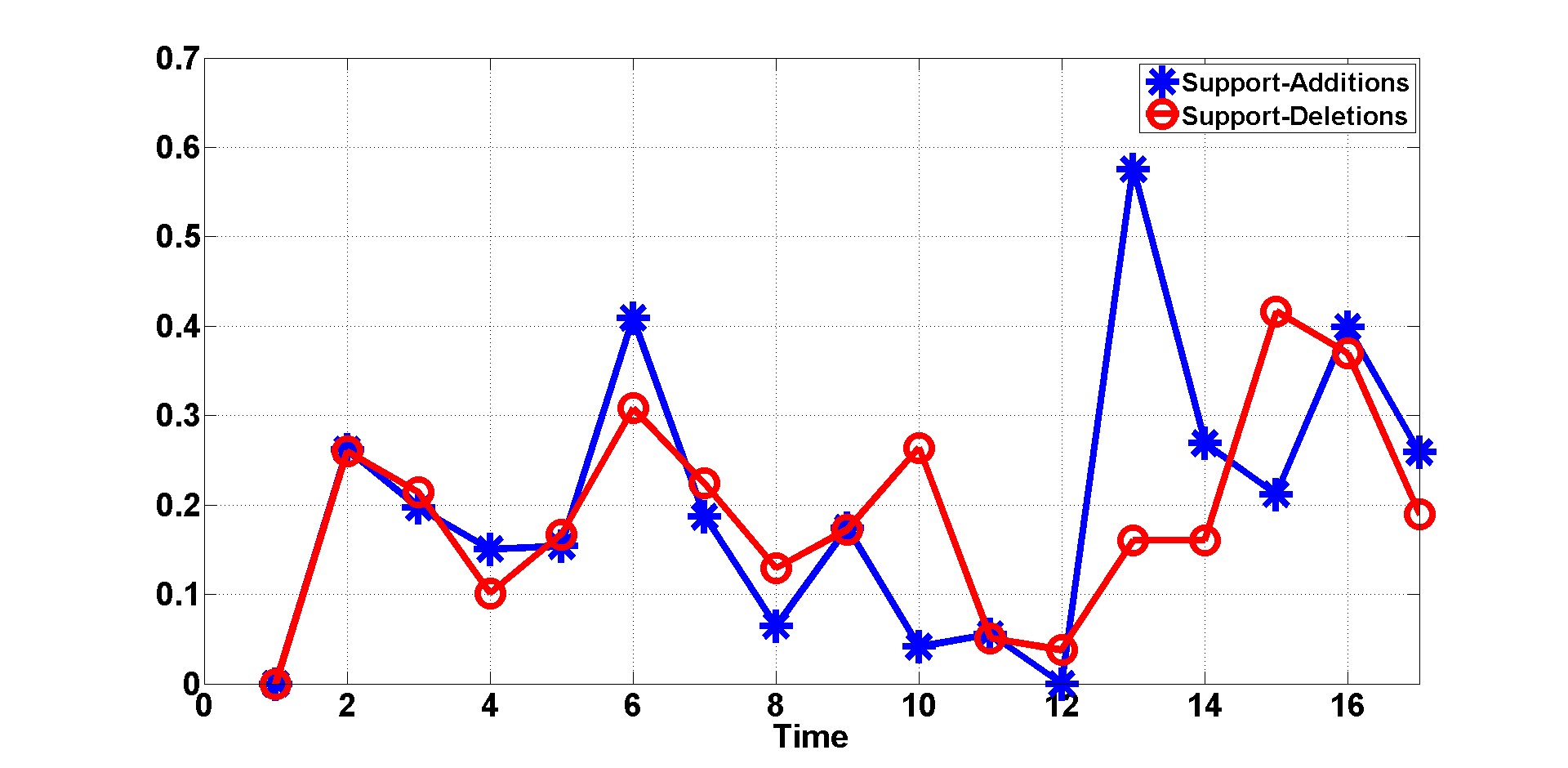}
\hspace{-0.4in}
&\includegraphics[height=50mm,width=50mm]{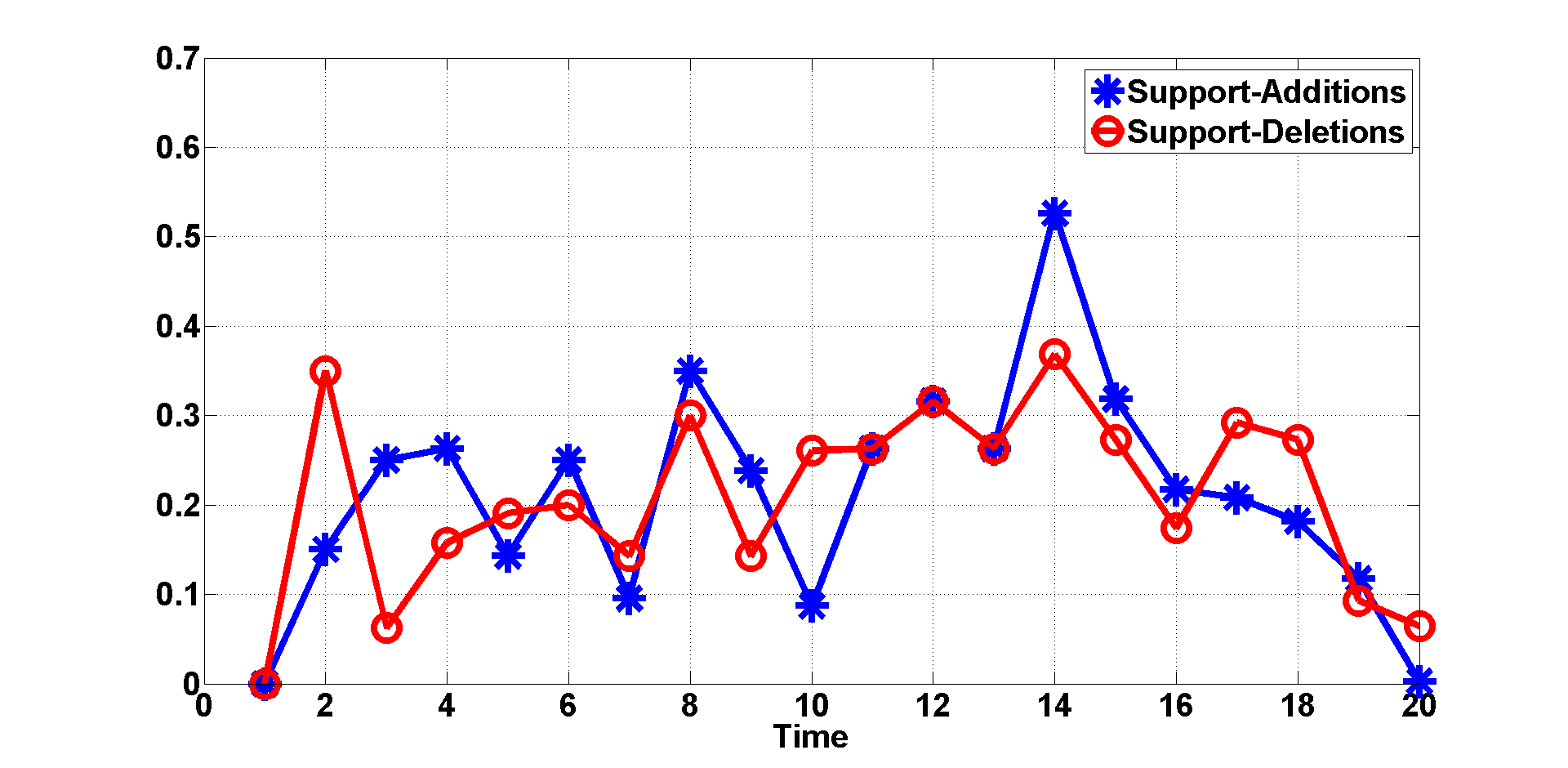}
\hspace{-0.4in}
&\includegraphics[height=50mm,width=50mm]{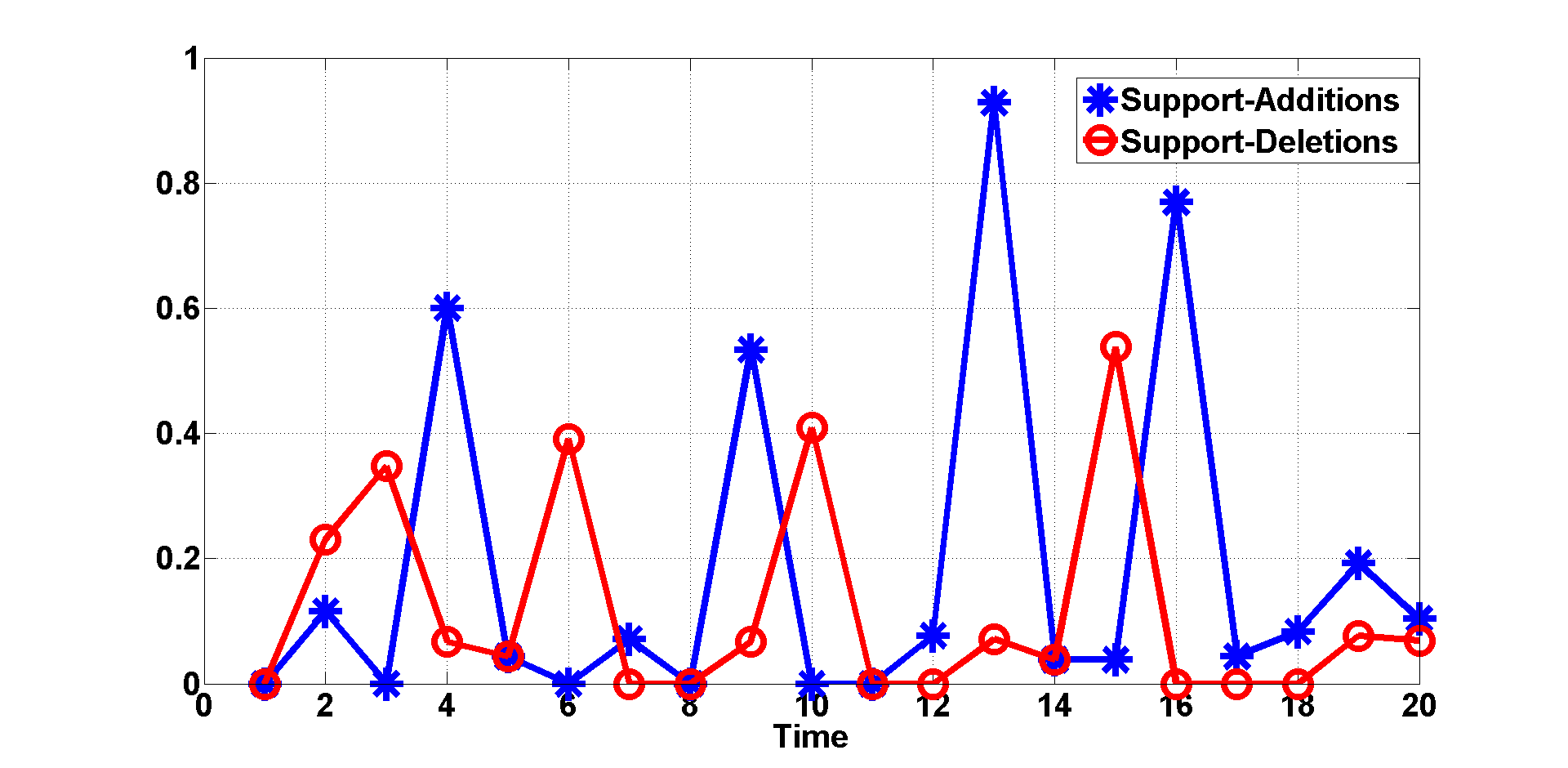}
\vspace{-0.15in}
\\
{\small outdoor-standing} & {\small outdoor-walking} & {\small indoor-walking}
\end{tabular}
}
}
\centerline{
\subfigure[entries in the support set at various times (shaded in black)]{
\label{blockplot}
\begin{tabular}[c]{ccc}
\includegraphics[height=70mm,width=70mm]{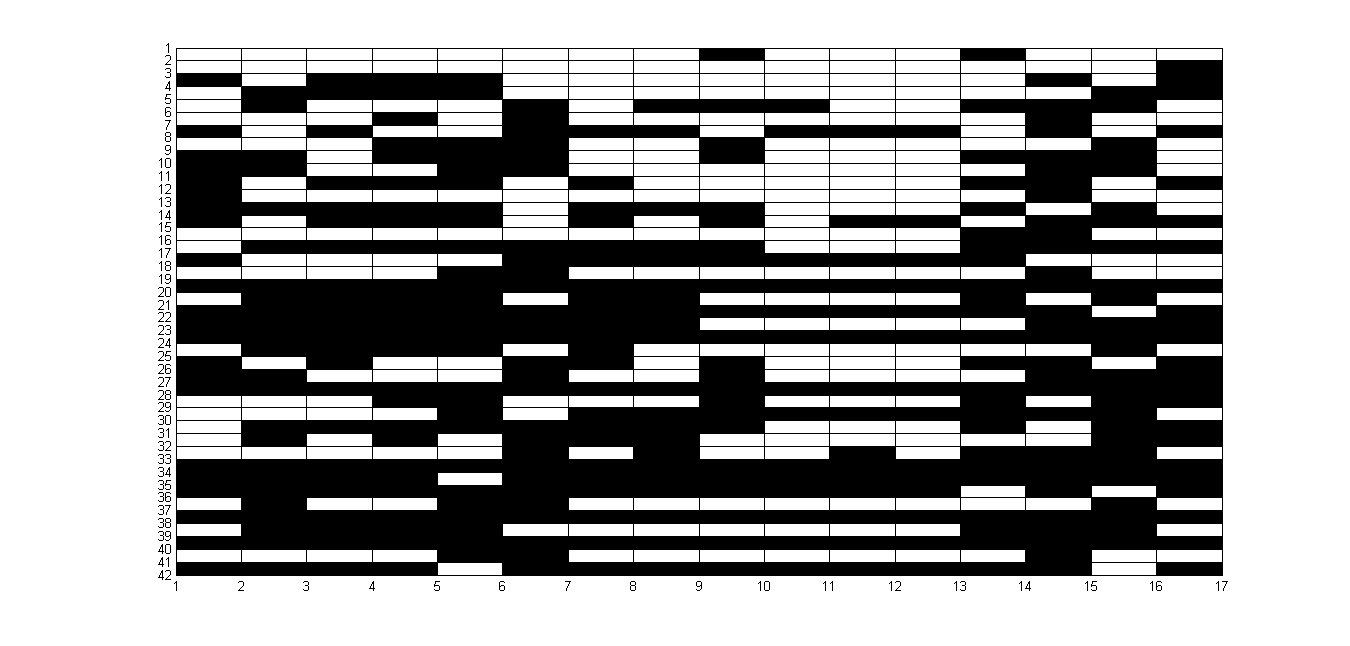}
\hspace{-0.4in}
&\includegraphics[height=70mm,width=70mm]{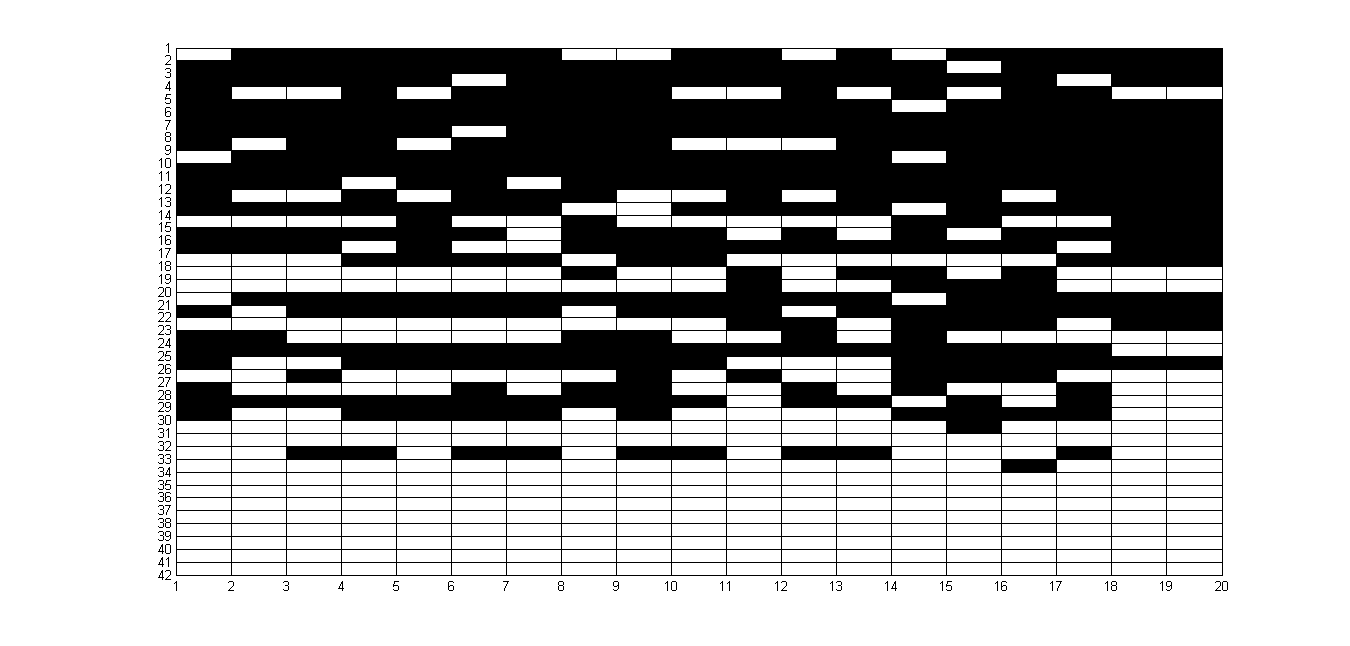}
\hspace{-0.4in}
&\includegraphics[height=70mm,width=70mm]{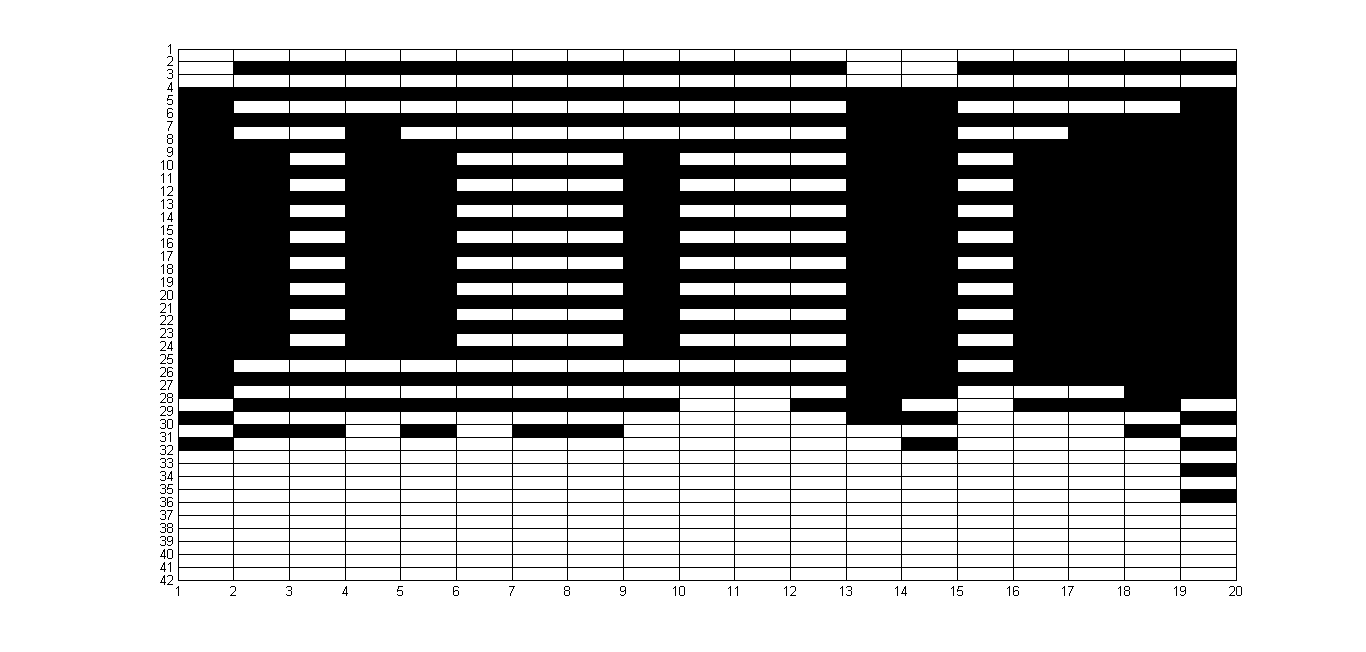}
\vspace{-0.25in}
\\
{\small outdoor-standing} & {\small outdoor-walking} & {\small indoor-walking}
\end{tabular}
}
}
\caption{{\small Sparsity and slow sparsity pattern change of the illumination image sequence for videos of Fig \ref{sample_illu} in the Legendre dictionary.
}}
\vspace{-0.15in}
\label{suppsize}
\end{figure*}

\begin{figure*}[!t]
\begin{center}
\includegraphics[height=60mm,width=100mm]{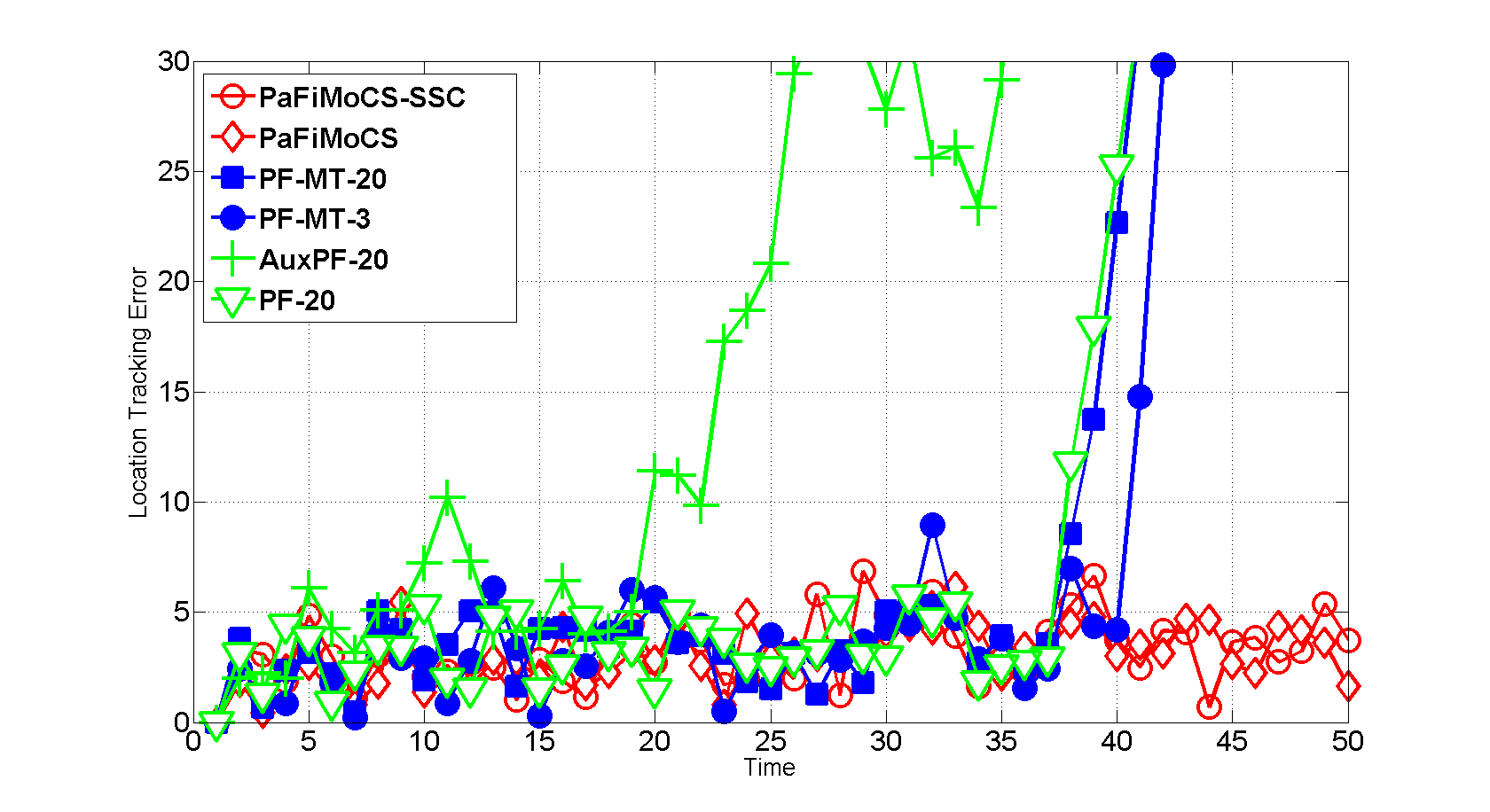}
\end{center}
\vspace{-0.3in}
\caption{{\small Location error plots for the outdoor-walking video. PaFiMoCS-SSC refers to PaFiMoCS-slow-support-change.
}}
\vspace{-0.2in}
\label{loc_error_video}
\end{figure*}

\begin{figure*}[!t]
\centering
\subfigure[tracking results using PaFiMoCS-slow-support-change]{
\begin{tabular}{ccccc}
\includegraphics[height=23mm,width=27mm]{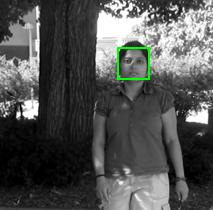}
&\includegraphics[height=23mm,width=27mm]{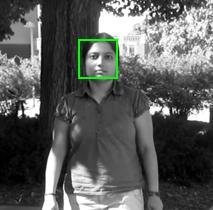}
&\includegraphics[height=23mm,width=27mm]{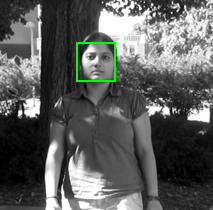}
&\includegraphics[height=23mm,width=27mm]{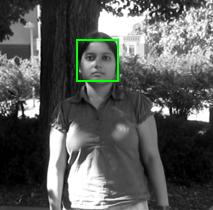}
&\includegraphics[height=23mm,width=27mm]{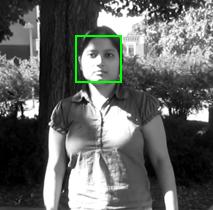}
\end{tabular}
}
\subfigure[tracking results using PaFiMoCS]{
\begin{tabular}{ccccc}
\includegraphics[height=23mm,width=27mm]{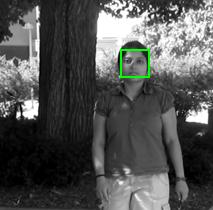}
&\includegraphics[height=23mm,width=27mm]{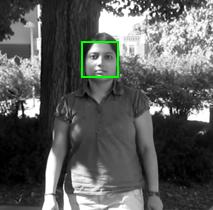}
&\includegraphics[height=23mm,width=27mm]{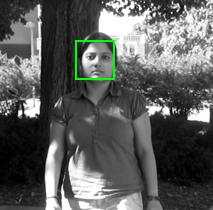}
&\includegraphics[height=23mm,width=27mm]{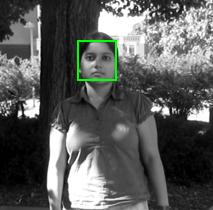}
&\includegraphics[height=23mm,width=27mm]{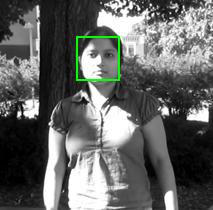}
\end{tabular}
}
\subfigure[tracking results using PF-MT-3]{
\begin{tabular}{ccccc}
\includegraphics[height=23mm,width=27mm]{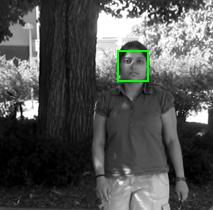}
&\includegraphics[height=23mm,width=27mm]{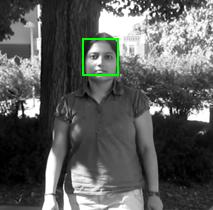}
&\includegraphics[height=23mm,width=27mm]{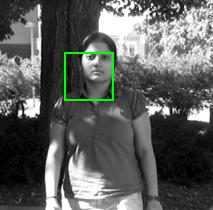}
&\includegraphics[height=23mm,width=27mm]{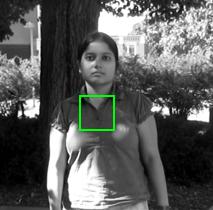}
&\includegraphics[height=23mm,width=27mm]{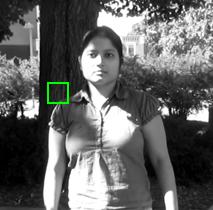}
\end{tabular}
}
\subfigure[tracking results using PF-MT-20]{
\begin{tabular}{ccccc}
\includegraphics[height=23mm,width=27mm]{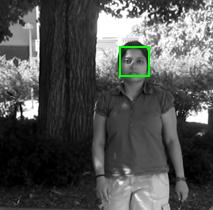}
&\includegraphics[height=23mm,width=27mm]{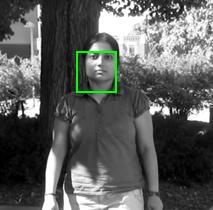}
&\includegraphics[height=23mm,width=27mm]{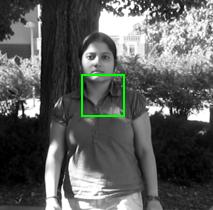}
&\includegraphics[height=23mm,width=27mm]{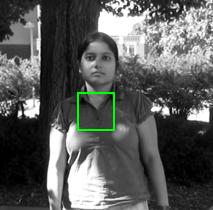}
&\includegraphics[height=23mm,width=27mm]{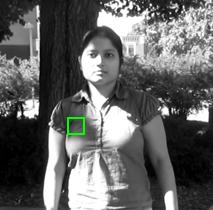}
\end{tabular}
}
\subfigure[tracking results using PF-Gordon-20]{
\begin{tabular}{ccccc}
\includegraphics[height=23mm,width=27mm]{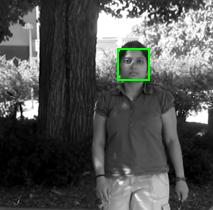}
&\includegraphics[height=23mm,width=27mm]{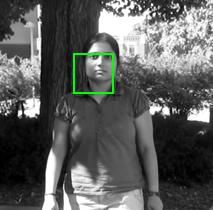}
&\includegraphics[height=23mm,width=27mm]{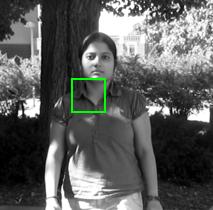}
&\includegraphics[height=23mm,width=27mm]{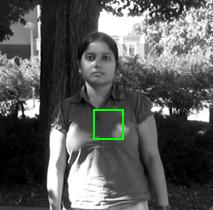}
&\includegraphics[height=23mm,width=27mm]{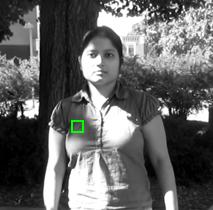}
\end{tabular}
}
\subfigure[tracking results using Auxiliary-PF-20]{
\begin{tabular}{ccccc}
\includegraphics[height=23mm,width=27mm]{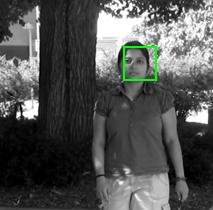}
&\includegraphics[height=23mm,width=27mm]{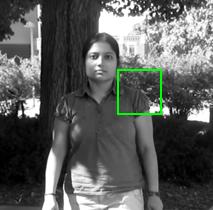}
&\includegraphics[height=23mm,width=27mm]{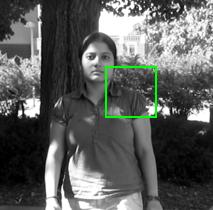}
&\includegraphics[height=23mm,width=27mm]{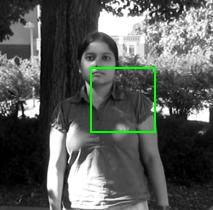}
&\includegraphics[height=23mm,width=27mm]{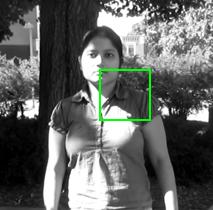}
\\
$t=15$ & $t=36$ & $t=41$ & $t=48$ & $t=56$
\end{tabular}
}
\caption{{\small
Tracking results for the outdoor-walking sequence. 
We show the comparison for frames 15, 36, 41, 48 and 56 respectively.
}}
\vspace{-0.2in}
\label{olivia_vid3}
\end{figure*}

However, since PF-MT does not exploit the sparsity or slow sparsity pattern change of $\Lambda_t$, with probability one, it results in a dense solution for $\Lambda_t$, i.e. the energy gets distributed among all components of $\Lambda_t$. This becomes a problem in applications where $\Lambda_t$ is indeed well approximated by a sparse vector with changing sparsity patterns. An alternative could be to assume (\ref{sysmod2}) on a selected fixed subset of $\Lambda_t$, i.e. fix $T_t = T_0$. For example, if $\Phi$ is the Fourier basis or a Legendre dictionary, one would pick the first few components as the set $T_0$. This was done in \cite{das_tip_11} for illumination. This approach works if most energy of $L_t$ does indeed lie in the lower frequency (or lower order Legendre) components, but fails if there are different types of high-frequency (higher order Legendre) spatial variations in $L_t$ over time\footnote{We show a few 2D Legendre polynomials (images of $P_k(i,j)$ defined in (\ref{def_phi}) for a few values of $k$) in Fig \ref{leg_illu}. As can be seen, higher order Legendre polynomials roughly correspond to higher frequency spatial variations of intensity.}. For many of the video sequences we experimented with for motion tracking across illumination change, this was indeed true, i.e. higher order Legendre polynomials were part of the Legendre support set of the illumination image [see Fig \ref{suppsize}], and as a result PF-MT implemented this way also failed [see Figs \ref{loc_error_video}, \ref{olivia_vid3}].



\Subsection{PaFiMoCS: Particle Filtered Modified-CS}
\label{pafimocs_algo2}
To address the above limitation, one can utilize the sparsity and slow sparsity pattern change of the large dimensional state vector, $L_t$, in a PF-MT type framework as follows. The key idea is to add a term motivated by Modified-CS \cite{isitmodcs} in the mode tracking optimization step and make corresponding changes in the other steps. We proceed as follows. We let $X_{t,s} = [U_t, T_t]$ and $X_{t,r} = \Lambda_t$. In the importance sampling step, we sample $U_t^i$ and $T_t^i$ from their state transition priors given in Sec \ref{statetrans}.  In the cost function that we minimize for the mode tracking step, we add a term of the form $\|\Lambda_{T^c}\|_1$ with $T = T_{t}^i$, i.e. it computes $\Lambda_t^i$ as the solution of $ \min_{\Lambda}  C(\Lambda) \ \text{where} \ $
\bea
C(\Lambda): =  -\log \text{OL}(U_t^i, \Phi \Lambda) + \beta \frac{\|(\Lambda - \Lambda_{t-1}^i)_{T_{t}^i}\|_2^2}{2 \sigma_l^2} + \gamma \|\Lambda_{(T_{t}^i)^c}\|_1
\label{pafimocs_cost}
\eea
where $\text{OL}(.)$ is defined in (\ref{ol_gen}). Solving (\ref{pafimocs_cost}) is a tractable approximation to trying to find the vector $\Lambda_t^i$ that is sparsest outside the set $T_t^i$ (i.e. the vector with the smallest number of new support additions to $T_t^i$) among all vectors $\Lambda$ that satisfy the observation model constraint (often referred to as the data constraint) and are ``close enough" to the previous estimate, $(\Lambda_{t-1}^i)_{T_t^i}$. Thus solving (\ref{pafimocs_cost}) ensures that the support of the solution, $\Lambda_t^i$, does not change too much w.r.t. the predicted support particle $T_t^i$. The larger the value of $\gamma$, the smaller will be the support change. Notice that (\ref{pafimocs_cost}) can also be interpreted as an adaptation of the regularized modified-CS idea which was originally introduced for linear problems with slow sparsity pattern and signal value change in \cite{regmodbpdn}.

\begin{algorithm}[!t]
\caption{\small{PaFiMoCS: Particle Filtered Modified-CS}}
\label{pafimocs_2_algo}
 Input: $Y_t$
\\ Output: $U_t^i, T_t^i, \Lambda_t^i, w_t^i$
\\ Parameters: (algorithm) $n_{pf}$, $\alpha, \gamma, \beta$, (model) $\Sigma_u, \sigma_l^2, p_a, p_r, f_Z(z)$
\\ For all $t \ge 0$ do
\ben
\item For each particle $i=1,2,\dots n_{pf}$: Importance sample $U_t$ from its state transition prior: $U_t^i \sim \n(0, \Sigma_u)$

\item For each particle $i=1,2,\dots n_{pf}$: Importance sample $T_t$ from its state transition prior: $T_t^i = T_{t-1}^i \cup A_t^i \setminus R_t^i$ where $A_t^i \sim \text{Ber}((T_{t-1}^i)^c,p_a)$ and $R_t^i \sim \text{Ber}(T_{t-1}^i,p_r)$.

\item For each particle $i=1,2,\dots n_{pf}$:  Mode track $\Lambda_t$ with imposing slow sparsity pattern change, i.e. compute $\Lambda_t^i$ as the solution of (\ref{pafimocs_cost}) with $\text{OL}(.)$ as defined in (\ref{ol_gen}).

\item For each particle $i=1,2,\dots n_{pf}$: Update $T_t^i$ as
\bea
T_t^i :=\{j: |(\Lambda_{t}^i)_j| > \alpha\} \nn
\eea

\item For each particle $i=1,2,\dots n_{pf}$:  Compute the weights as follows
$$w_t^i \propto w_{t-1}^i \ \text{OL}(U_t^i, \Phi \Lambda_t^i) \ \text{STP}(\Lambda_t^i ;\Lambda_{t-1}^i, T_{t}^i) $$ 
where $\text{OL}(.)$ is defined in (\ref{ol_gen}) and $\text{STP}(\Lambda_t^i ;\Lambda_{t-1}^i, T_{t}^i)$ is defined in (\ref{stp_lam}).

\item Resample and reset weights to $1/n_{pf}$. Increment $t$ and go to step 1.
\een
\end{algorithm}

A second change that we have w.r.t. the original PF-MT idea is that we threshold on $\Lambda_t^i$ in order to get an updated estimate of the support $T_t$. We compute it as $T_t^i :=\{j: |(\Lambda_{t}^i)_j| > \alpha\}$ for a small nonzero threshold $\alpha$. Finally, as in PF-MT, the weighting step multiplies the previous weight by a product of the observation likelihood and the state transition prior of $X_{t,r} = \Lambda_t$.
We summarize the resulting algorithm in Algorithm \ref{pafimocs_2_algo}. We refer to it as {\em Particle Filtered Modified-CS (PaFiMoCS)}. In situations where $\Lambda_t$ is indeed well approximated by a sparse vector with changing sparsity patterns, this significantly improves tracking performance as compared to PF-MT as well as other PF algorithms. We demonstrate this for the illumination-motion tracking problem in Sec. \ref{expts}.

We should point out here, that from the Bayesian perspective, in (\ref{pafimocs_cost}), the multiplier $\beta$ should be one and $\gamma$ should be chosen by assuming that the elements of $(\Lambda_t)_{T^c}$ are i.i.d. Laplace distributed and the Laplace parameter can be estimated by maximum likelihood estimation. For details, see \cite[Appendix C]{isitmodcs}. However, it is well known that in solving sparse recovery problems, the weights given by the Bayesian perspective are not always the best ones to use. A more practically useful approach to selecting these is explained in \cite[Section V-B]{regmodbpdn}.

\Subsection{PaFiMoCS-slow-support-change: PaFiMoCS for large sized problems with slow support changes}
\label{pafimocs_algo1}
For certain problems with very large sized spatial signals, $L_t$, the support size of its sparse coefficients vector, $\Lambda_t$, can also be very large. In these situations, if we keep $T_t$ as part of the importance sampling state $X_{t,s}$, it will require a very large number of particles, thus making the algorithm impractical.  However, if the support changes slowly enough, then we can include $T_t$ as part of $X_{t,r}$, i.e. we let $X_{t,s} = U_t$ and $X_{t,r} = [T_t, \Lambda_t]$. With this, the mode tracking step would ideally have to compute $\Lambda_t^i$, $A_t^i$ and $R_t^i$ by solving\footnote{This follows by including the negative logarithm of state transition prior of $T_t$ given in (\ref{stp_T}) in the cost.}
\bea
&& \min_{\Lambda,A,R}  C(\Lambda,A,R) \ \text{where} \nn \\
&& C(\Lambda,A,R): =  -\log \text{OL}(U_t^i, \Phi \Lambda) + \beta \frac{\|(\Lambda - \Lambda_{t-1}^i)_{T_{t-1}^i}\|_2^2}{2 \sigma_l^2} - |A| \log \frac{p_a}{1-p_a} - |R| \log \frac{p_r}{1-p_r} \ \ \ \
\label{pafimocs_ssc_cost}
\eea
and setting $T_t^i = T_{t-1}^i \cup A_t^i \setminus R_t^i$. But the above minimization will require a brute force approach of checking all possible sets $A$ and $R$ and will thus have complexity that is exponential in the support change size. Thus, it cannot be solved in any reasonable time. However, we can instead compute $\Lambda_t^i$ by solving $\min_{\Lambda}  C(\Lambda) \ \text{where} $ 
\bea
&& C(\Lambda): =  -\log \text{OL}(U_t^i, \Phi \Lambda) + \beta \frac{\|(\Lambda - \Lambda_{t-1}^i)_{T_{t-1}^i}\|_2^2}{2 \sigma_l^2} + \gamma \|\Lambda_{(T_{t-1}^i)^c}\|_1
\label{pafimocs_cost_large_supp}
\eea
and then threshold on $\Lambda_t^i$ to get the current support particle $T_t^i$. Since $p_a < 0.5$ and $p_r < 0.5$, the last two terms of (\ref{pafimocs_ssc_cost}) are increasing functions of $|A|$ and $|R|$. If the last term of (\ref{pafimocs_ssc_cost}) were ignored, doing the above can be interpreted as its convex relaxation: it helps to find the vector $\Lambda_t^i$ with the smallest number of support additions to $T_{t-1}^i$, i.e. the smallest $|A_t^i|$, while also keeping the first two terms small. If the last term of (\ref{pafimocs_ssc_cost}) is not ignored, then it is not clear what its convex relaxation would be.

Since the support set $T_t$ is now a part of $X_{t,r}$, we also need to include a term proportional to its state transition prior in the weighting step, i.e. we need to also multiply by (\ref{stp_T}) in the weighting step. We summarize the resulting algorithm in Algorithm \ref{pafimocs_1_algo}. We refer to it as {\em PaFiMoCS-slow-support-change}. We should note that this works only as long as the support changes at all times are slow enough. 

\begin{algorithm}
\caption{\small{PaFiMoCS-slow-support-change: PaFiMoCS for slow support changes}}
\label{pafimocs_1_algo}
 Input: $Y_t$, \  Output: $U_t^i, T_t^i, \Lambda_t^i, w_t^i$
\\ Parameters: (algorithm) $\alpha, \gamma, \beta$, (model) $\Sigma_u, \sigma_l^2, p_a, p_r, f_Z(z)$
\\ For all $t \ge 0$ do
\ben
\item For each particle $i=1,2,\dots n_{pf}$: Importance sample $U_t$ from its prior: $U_t^i \sim \n(0, \Sigma_u)$

\item For each particle $i=1,2,\dots n_{pf}$:  Mode track $\Lambda_t$, $T_t$ with imposing slow sparsity pattern change, i.e. compute $\Lambda_t^i$ as the solution of (\ref{pafimocs_cost_large_supp})  with $\text{OL}(.)$ as defined in (\ref{ol_gen}).  Compute $T_t^i$ by thresholding on $\Lambda_t^i$, i.e. compute
\bea
T_t^i :=\{j: |(\Lambda_{t}^i)_j| > \alpha\} \nn
\eea

\item For each particle $i=1,2,\dots n_{pf}$:  Compute the weights as follows
$$w_t^i \propto w_{t-1}^i \text{OL}(U_t^i, \Phi \Lambda_t^i) \ \text{STP}(\Lambda_t^i ;\Lambda_{t-1}^i, T_{t}^i) \ \text{STP}(T_t^i; T_{t-1}^i) $$ 
where $\text{OL}(.)$ is defined in (\ref{ol_gen}),  $\text{STP}(\Lambda_t^i ;\Lambda_{t-1}^i, T_{t}^i)$ is defined in (\ref{stp_lam}) and $\text{STP}(T_t^i; T_{t-1}^i)$ is defined in (\ref{stp_T}).

\item Resample and reset weights to $1/n_{pf}$. Increment $t$ and go to step 1.
\een
\end{algorithm}

\Subsection{Convexity of the Mode Tracking Cost Function}
\label{convexity}
Consider the cost function $C(\Lambda)$ that we need to minimize in either of the above PaFiMoCS algorithms (Algorithm \ref{pafimocs_1_algo} or \ref{pafimocs_2_algo}). If this is convex, any minimizer is a global minimizer. Also, there exist a large number of efficient algorithms, such as interior point methods as well as other more recent efficient algorithms, for minimizing convex functions. There are also multiple software packages such as CVX (CVX: Matlab software for disciplined convex programming, \url{http://cvxr.com/cvx}), that contain efficient implementations of these algorithms. If the cost function is not convex, as long as it is differentiable, one can still try to use the convex solvers, but one will only find a local minimizer that is closest to the initial guess provided.

One simple set of sufficient conditions for $C(\Lambda)$ to be convex, that are also often satisfied in practice, are as follows.
\ben
\item $Z_t$ is Gaussian distributed, i.e. $f_Z(z)$ is a Gaussian PDF and
\item The observation model is such that $g(Y_t,U_t^i, \Phi \Lambda)$ is an affine function of $\Lambda$, i.e.
\bea
g(Y_t,U_t^i, \Phi \Lambda) =  g_{u,1}(Y_t,U_t^i) \Phi \Lambda + g_{u,2}(Y_t,U_t^i)
\label{affine_func}
\eea
\een
The illumination-motion tracking problem that we describe in Sec \ref{illum} is an example of a problem where the above assumptions hold.

A convex cost function ensures that any minimizer that we find is a global minimizer. Also, under certain extra assumptions, the minimizer will be unique. For example, if the above two assumptions hold, and if the matrix $B:= g_{u,1}(Y_t,U_t^i)\Phi$ satisfies the conditions imposed on the measurement matrix in \cite[Theorem 1]{regmodbpdn}, the minimizer will be unique.

\Subsection{Computational Cost and Parallel Implementation}
\label{parallel}
Compared with most other PF algorithms (except PF-MT, RB-PF or some others), PaFiMoCS has the advantage that it requires much fewer number of particles for tracking a large dimensional spatial signal and other states over time. However, like PF-MT, PaFiMoCS also needs to solve a convex optimization problem for each particle. This can be make its implementation speed quite slow. However, notice that, like PF-MT, PaFiMoCS is ideally suited for a parallel implementation since the convex optimization needs to be solved for each particle independently of all the others. A parallel implementation would easily enable PaFiMoCS to run in real-time. 

\Subsection{Dealing with Outliers}
\label{outliers}
Outliers in the observation noise occasionally occur in almost all practical tracking problems. An outlier is a very large value of the noise that occurs with very small probability. However since it is a very large value, even one outlier occurrence can cause the tracker to lose track and thus any practical tracker needs to be able to deal with outliers.  In computer vision, the most common reason for an outlier is an occlusion.

There are two commonly used ways to model outliers. The traditional approach \cite{condensation} is to let the observation noise $Z_t$ be a Gaussian mixture with one large variance Gaussian and one small variance one, i.e.
$$Z_t \stackrel{\text{i.i.d.}}{\sim} f_Z(z) = \prod_{i=1}^{n} [ (1-p_{\text{out}})  \n(z_i; 0, \sigma^2) + p_{\text{out}} \n(z_i; 0, \sigma_{\text{out}}^2)] $$
with $\sigma_{\text{out}}^2 \gg \sigma^2$, e.g. $\sigma_{\text{out}}^2 = 1000 \sigma^2$ and with the outlier probability $p_{\text{out}}$ being very small.
Models similar to this one were used in \cite{pfmtpap} and in \cite{das_tip_11}. However, one main disadvantage of using this is that it makes the cost function that we need to minimize non-convex. 

A more recent and also more efficient approach to model outliers is treat them as sparse vectors \cite{error_correction_PCP_l1}. Thus, one lets
$$Z_t = Z_{t,g} + O_t$$
where $Z_{t,g} \sim  \n(0, \sigma^2 I)$ is the usual Gaussian noise while $O_t$ is an unknown sparse vector. With this model, in the mode tracking step of Algorithm \ref{pafimocs_1_algo} or Algorithm \ref{pafimocs_2_algo}, one solves
\bea
&& \min_{\Lambda,O}  C(\Lambda,O), \ \text{where}  \nn \\
&& C(\Lambda,O) = \frac{\|g(Y_t,U_t^i,\Phi \Lambda) - O\|_2^2}{2\sigma^2} + \beta \frac{\|(\Lambda - \Lambda_{t-1}^i)_T\|_2^2}{2 \sigma_l^2} + \gamma \|\Lambda_{T^c}\|_1 + \gamma' \|O\|_1 \nn
\eea
Notice that if $g(Y_t,U_t^i, \Phi \Lambda])$ is an affine function of $\Lambda$, i.e. if it satisfies (\ref{affine_func}), then this cost function is still convex. 















\Section{Visual Tracking across Spatially varying Illumination Changes}
\label{illum}
In this section, we show how visual tracking across spatially varying illumination change is an example of the general problem studied here and how to design PaFiMoCS for it. We give the observation model for this problem in Sec \ref{obsmod_sec} below, followed by the state transition model in Sec \ref{sysmod_sec}. In Sec \ref{supp_change_sec}, we demonstrate that for many video sequences, the illumination image is indeed sparse in the Legendre dictionary and most of the time its sparsity pattern does change slowly over time. The PaFiMoCS and PaFiMoCS-support algorithms for this problem are summarized in Sec \ref{pafimocs_illum}.

\Subsection{Observation Model}
\label{obsmod_sec}

The state in this case consists of the $n_u \times 1$ motion state, $U_t$, which is the small dimensional part, and the $n_l \times 1$ illumination ``image" (written as 1-D vector), $L_t$.
In this paper, we use a template-based tracking framework, similar to the one in \cite{das_tip_11}, with a simple three-dimensional motion model, that only models x-y translation and scale, i.e. $U_t =[u_t^x, u_t^y, s_t]^\top$ where $s_t$ refers to scale and $u_t^x$ and $u_t^y$ refer to x and y translation. Thus $n_u = 3$. 
The illumination image is represented in the Legendre dictionary. Thus, our final state vector is $X_t = [U_t^\top, T_t^\top, \Lambda_t^\top]^\top$ where $U_t$ is the $n_u \times 1$ motion state, $\Lambda_t$ is the  $n_\lambda \times 1$ Legendre coefficients' vector of illumination and $T_t$ is the support set of $\Lambda_t$. 

The observation model used in this work is taken from \cite{das_tip_11}. We repeat it here for completeness.  The initial template is denoted by $I_0$. We use $\text{ROI}(U_t)$ to denote the region-of-interest (\text{ROI}) in the current image, i.e. it contains the pixel locations of the template in the current frame. The total number of pixels in $I_0$ is $n_l$ and thus the size of the set $\text{ROI}(U_t)$ is also $n_l$. The total number of pixels in the entire image, $Y_t$, is $m > n_l$. At time $t$, $\text{ROI}(U_t)$ is obtained by scaling and translating the pixel locations of the original template $I_0$, i.e. by using (\ref{roi}). The illumination image is represented using the Legendre dictionary defined in (\ref{def_phi}).
The pixel intensities of the pixels in the \text{ROI} are modeled as being equal to those of the initial template plus the initial template scaled pixel-wise by the illumination image plus some randomness that is modeled as Gaussian noise (in the absence of occlusion). The pixels outside the \text{ROI}, i.e. those in $\text{ROI}(U_t)^c$, are assumed to be due to clutter and we simply model them as being i.i.d. uniformly distributed between zero and 255. This model can be mathematically specified as follows. 
\bea
Y_t(\text{ROI}(U_t)) \se \text{vec}(I_0) + (\text{vec}(I_0) \varodot L_t) + Z_t = \text{vec}(I_0) + \Phi \Lambda_t + Z_t, \ Z_t \sim \n(0, \sigma_o^2 I), \nn \\
Y_t(\text{ROI}(U_t)^c) \se Z_{t,c}, \ (Z_{t,c})_i  \stackrel{\text{i.i.d.}}{\sim}  \text{Unif}(0,255)   
\label{yt_mod}
\eea
where ``i.i.d." means i.i.d. spatially and over time;
\bea
\label{roi}
{\text{ROI}}(U_t) &:=& \{ \text{round}({\bf J_1}U_t + {\bf 1}\bar{i}_0), \text{round}({\bf J_2}U_t + {\bf 1} \bar{j}_0) \}, \ \text{where}  \nn \\
{\bf J_1} &:=& [{\bf 1} \ {\bf 0} \ (\underline{i_0} - {\bf 1} {\bar{i}_0})], \ {\bf J_2} := [{\bf 0} \ {\bf 1} \ (\underline{j_0} - {\bf 1}{\bar{j}_0})],
\eea
$\underline{i_0}$ and $\underline{j_0}$ contain the x coordinates and the y coordinates respectively of the initial template $I_0$, ${\bar{i}_0} := \text{mean}(\underline{i_0}) = \frac{1}{M}\sum_{k=1}^M [\underline{i_0}]_k$, ${\bar{j}_0} := \text{mean}(\underline{j_0}) = \frac{1}{M}\sum_{k=1}^M [\underline{j_0}]_k$ denote the x and y location of the centroid of $I_0$ (since the template is of size $n_l$, $\underline{i_0}$ and $\underline{j_0}$ are also $n_l$ length vectors and ${\bf J_1}, {\bf J_2}$ are $n_l \times 3$ matrices); and
\bea
\Phi \sdefn [\text{vec}(I_0 \varodot P_0),....,\text{vec}(I_0 \varodot P_{2d})], \ \text{where} \nn \\
P_k(i,j) \se \begin{cases}
1 \; \; \text{if} \; \; k = 0 \\
p_{\frac{k+1}{2}}(i) \; \; \text{if} \; \; k = 1, 3, 5, \dots (2d-1) \\
p_{\frac{k}{2}}(j) \; \; \text{if} \; \; k = 2, 4, 6, \dots 2d
\end{cases}
\label{def_phi}
\eea
and $p_k(.)$ is the Legendre polynomial of $k^{th}$ order. Thus, $\Phi$ is an $n_l \times n_\lambda$ matrix with $n_\lambda = 2d+1$.
We show images of $P_k(i,j)$ for a few values of $k$ in Fig \ref{leg_illu}. As can be seen, higher order Legendre polynomials roughly correspond to higher frequency spatial variations of intensity.
In our experiments, we used $d=20$, so that $n_\lambda = 41$, while the template size, $n_l$, was much larger. Thus, $\Phi$ was a tall matrix. 

The above model is of the form (\ref{obsmod_gen}) with $g(Y_t,U_t,\Phi \Lambda_t) = \vect{Y_t({\text{ROI}}(U_t)) - \text{vec}(I_0) - \Phi\Lambda_t}{Y_t(\text{ROI}(U_t)^c)}$. Thus, $\text{OL}(.)$ is of the form (\ref{ol_gen}) and can be explicitly written as
\bea
\label{ol_gen_illum}
OL(U_t,\Phi \Lambda_t) =  {\cal N}([Y_t(\text{ROI}(U_t))]- \text{vec}(I_0)- \Phi \Lambda_t; 0, \sigma_o^2 I) \ \left(\frac{1}{255} \right)^{m - n_l}
\eea
Recall that $m$ is the size of the entire image $Y_t$ while $n_l$ is the size of the template.



\Subsection{System Model}
\label{sysmod_sec}
In the absence of any extra information about the motion, we assume a simple linear Gaussian random walk model on the motion state $U_t$, i.e. we assume that $U_t$ satisfies (\ref{sysmod3}) with $\Sigma_u$ being a diagonal matrix. As we show in Sec \ref{supp_change_sec} below, the  Legendre coefficients vector for illumination, $\Lambda_t$, is an approximately sparse vector with support that usually changes slowly over time. Hence, the models given in Sec \ref{problem} apply for $\Lambda_t$ as well: its support, $T_t$, satisfies (\ref{sysmod1}) and $\Lambda_t$ satisfies (\ref{sysmod2}). Thus, the corresponding state transition priors are also the same as those given there.

\Subsection{Verifying Illumination Image Sparsity and Slow Sparsity Pattern Change}
\label{supp_change_sec}

We used the video sequences shown in Fig \ref{sample_illu} to study the sparsity and sparsity pattern change of illumination images over time in the Legendre dictionary and in the Fourier basis. This was done as follows.
\ben
\item We took a 20 frame sub-sequence of each video. In each frame, a rectangle was hand-marked around the person's face. The rectangular region from the first frame served as the initial template $I_0$.

\item Let $I_t$ denote the rectangular region containing the person's face at time $t$. This was first resized to make it the same size as $I_0$. We then computed the maximum likelihood estimate of $\Lambda_t$ for the model $\text{vec}(I_t) = \text{vec}(I_0) + \Phi \Lambda_t + Z_t$ where $Z_t \sim \n(0, \sigma_l^2 I)$ as $\Lambda_t = (\Phi^\top \Phi)^{-1}\Phi^\top \text{vec}(I_t-I_0)$.

\item The vector ${\Lambda}_t$ computed above is not exactly sparse, but approximately so. We let $T_t$ be its {\em $99\%$-energy support} \cite{isitmodcs}, i.e. $T_t: = \{j: (\Lambda_t)_j^2 > \alpha\}$ where $\alpha$ is the largest real number for which $T_t$ contains at least $99\%$ of the signal energy. This is computed as follows: sort elements of $\Lambda_t$ in decreasing order of magnitude and keep adding elements to $T_t$ until $\sum_{j \in T_t} (\Lambda_t)_j^2 \ge 0.99 \sum_{j=1}^{n_\lambda} (\Lambda_t)_j^2$. We refer to $\alpha$ as the {\em 99\%-energy threshold.}


\item  In Fig \ref{suppsizeplot}, we plot the support size normalized by the length of $\Lambda_t$, i.e. $\frac{|T_t|}{n_\lambda}$ for all the three sequences.

\item In Fig \ref{suppchangeplot}, we plot the number of additions to, and removals from, the support normalized by the support size, i.e. we plot $\frac{|T_t \setminus T_{t-1}|}{|T_t|}$ and $\frac{|T_{t-1} \setminus T_t|}{|T_t|}$.

\item  In Fig \ref{blockplot}, we show which Legendre polynomials are contained in the support set $T_t$ at various times $t$ in each of the video sequences. We shade in black the squares corresponding to indices that are contained in $T_t$, while leaving blank the indices that are not in $T_t$.

\een

\begin{figure*}
\centerline{
\subfigure[support size]{
\begin{tabular}{c}
\includegraphics[height=45mm,width=50mm]{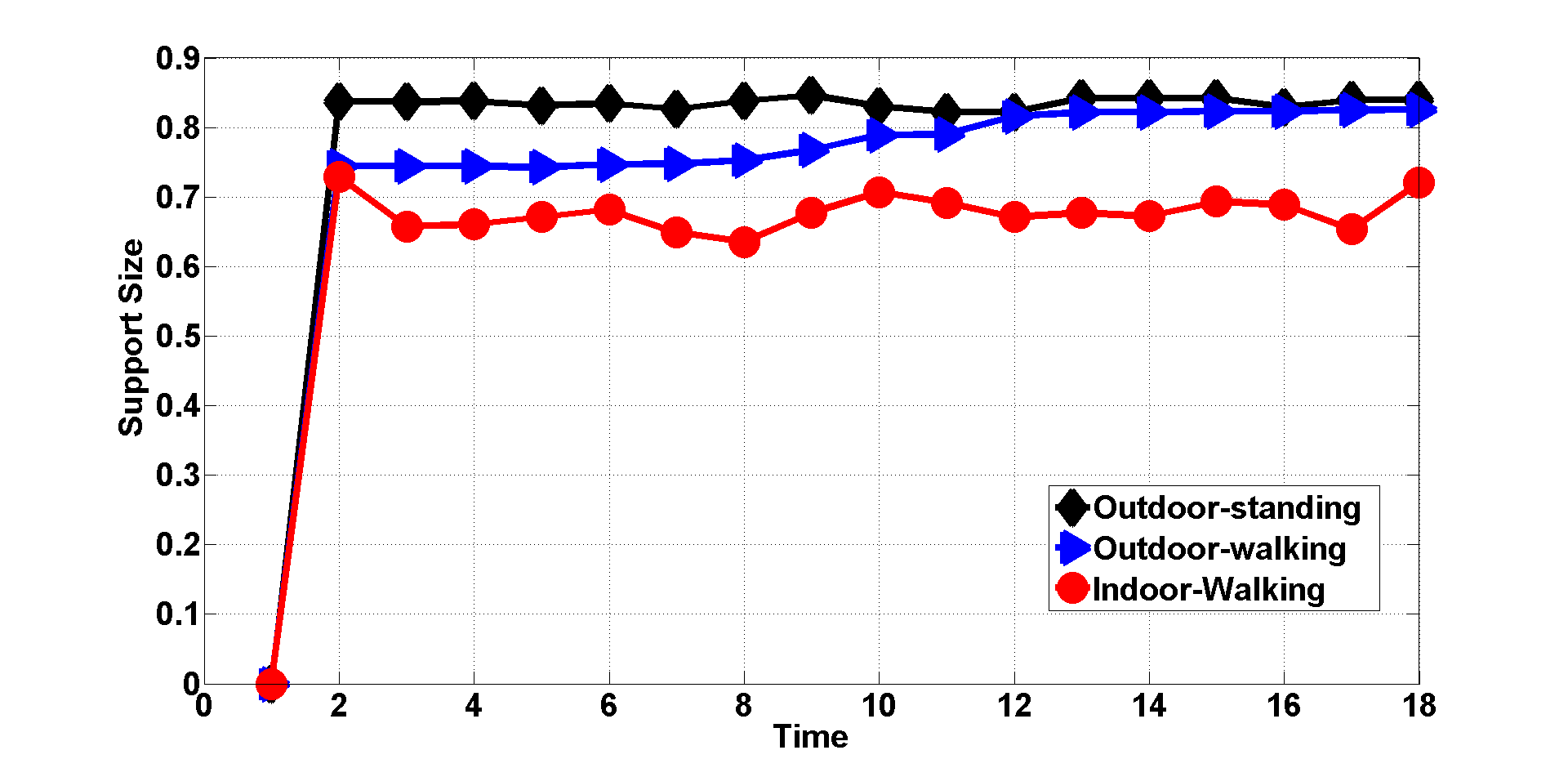}
\end{tabular}
}
\subfigure[support change sizes]{
\begin{tabular}{ccc}
\includegraphics[height=40mm,width=42mm]{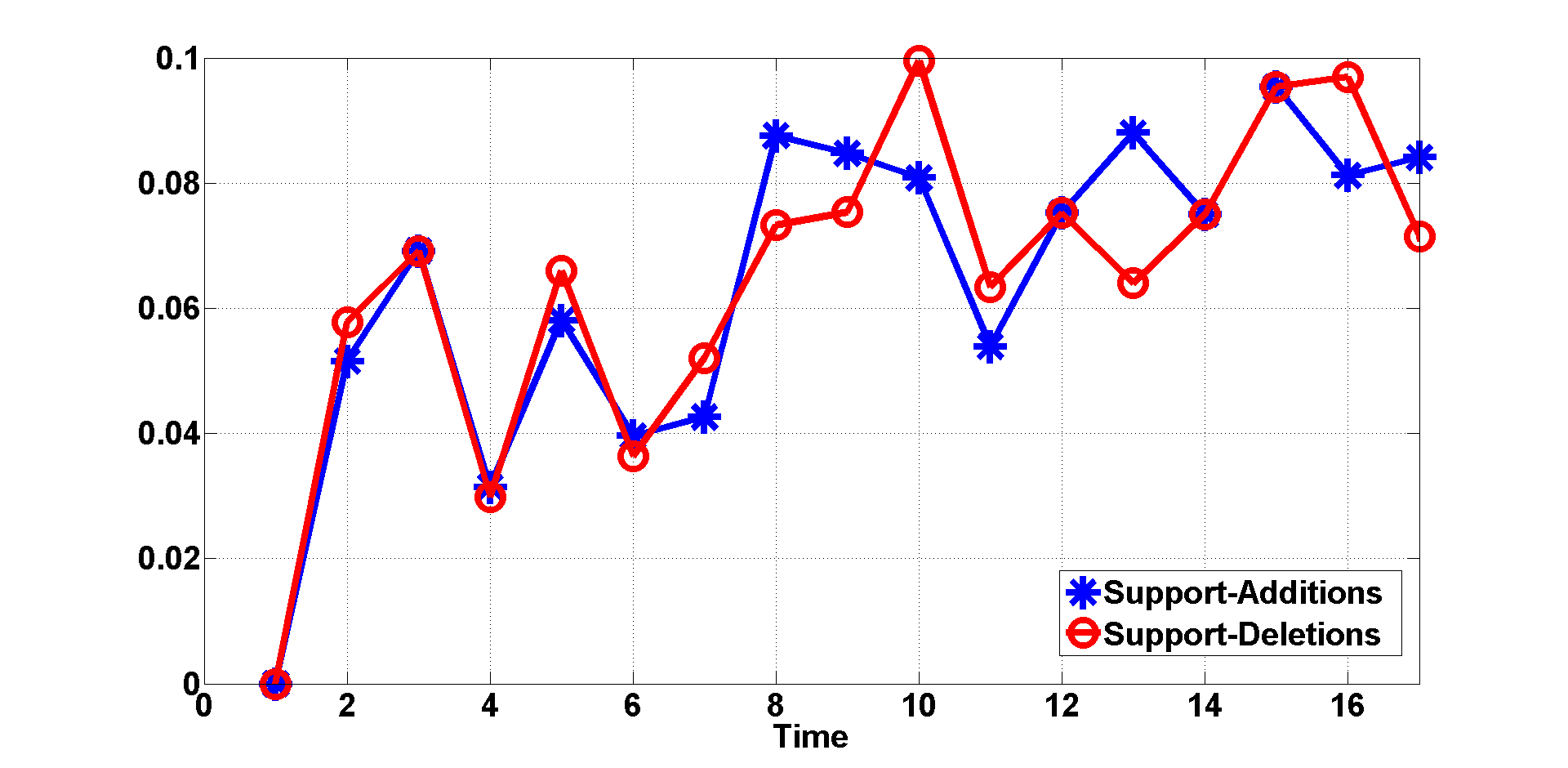}
&\includegraphics[height=40mm,width=42mm]{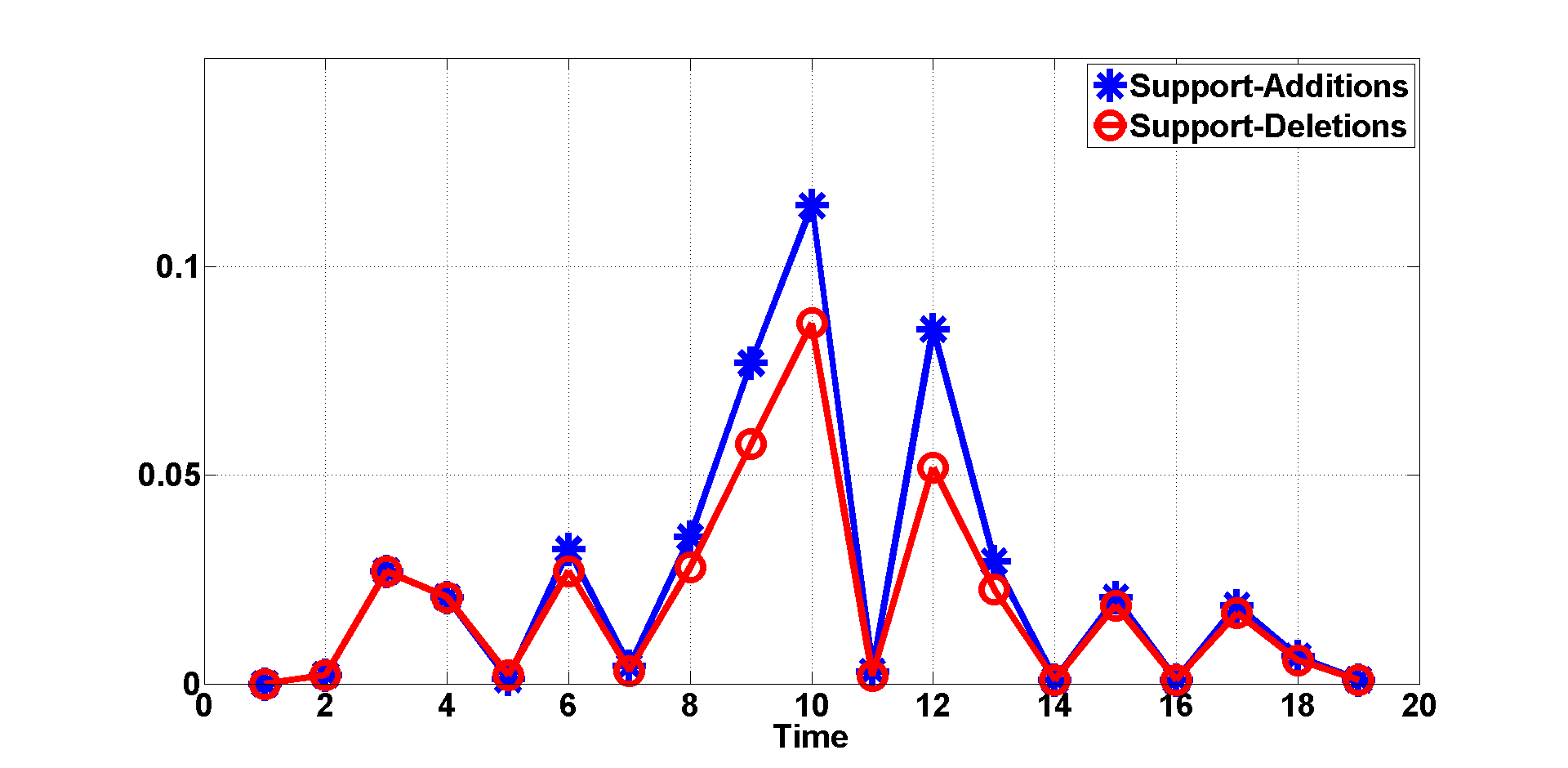}
&\includegraphics[height=40mm,width=42mm]{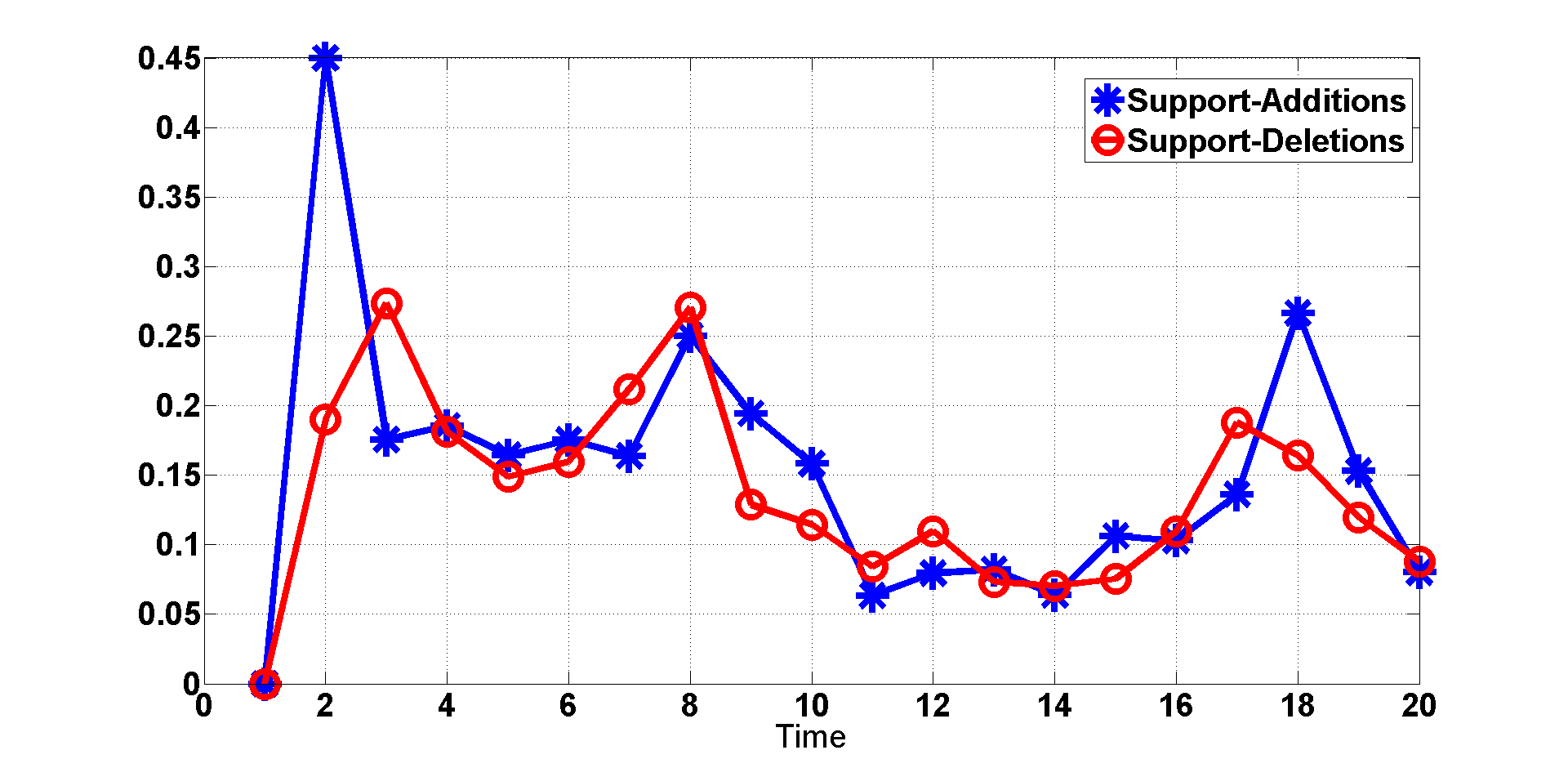}
\\
{\small outdoor-standing} & {\small outdoor-walking} & {\small indoor-walking}
\end{tabular}
}
}
\caption{{\small Support size and support change sizes of the illumination image sequence for the three videos of Fig \ref{sample_illu} in the Fourier basis.}}
\label{suppsize_Fourier}
\end{figure*}

We first did the above for the three video sequences shown in Fig \ref{sample_illu}, with using the Legendre dictionary, i.e. $\Phi$ was computed using (\ref{def_phi}) with $d=20$. The first (outdoor-standing) is of a person standing under a tree on a very windy day. Variable lighting falls on different parts of her face because leaves block some of it. As the leaves move, this pattern changes with time. The second (outdoor-walking) is of a person walking under a tree with variable amounts of light falling on various parts of her face as she moves under the moving leaves. The third (indoor-walking) has a person walking along a corridor across a window with varying illumination coming through the window.  The lighting is significantly more when he gets near the window and again lesser when he moves away from the window. Moreover, since the window is on the side of the person, the amount of light falling on different parts of his face is quite different (higher frequency spatial variations of illumination exist) and also changes with time. 

As can be seen from Fig \ref{suppsizeplot}, for all three sequences, the support size is between 30-60\% of the length of $\Lambda_t$, $n_\lambda$. Thus, $\Lambda_t$ is indeed approximately sparse.  Moreover, except at a few time instants, the number of support changes (additions or deletions) is usually under 35\% of the support size. For the indoor-walking sequence, at certain times (when the person moves towards the window from a darker region of the corridor or vice versa), the number of support changes is much larger than this.

\begin{figure*}[!t]
\centering
\subfigure[tracking results using PaFiMoCS-slow-support-change]{
\begin{tabular}{ccccc}
\includegraphics[height=20mm,width=24mm]{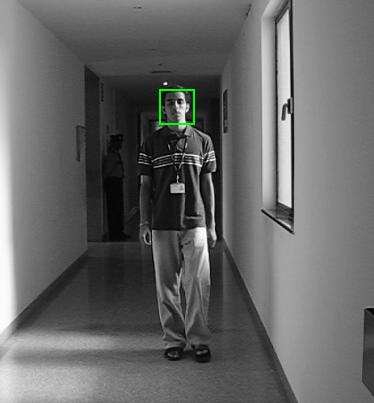}
&\includegraphics[height=20mm,width=24mm]{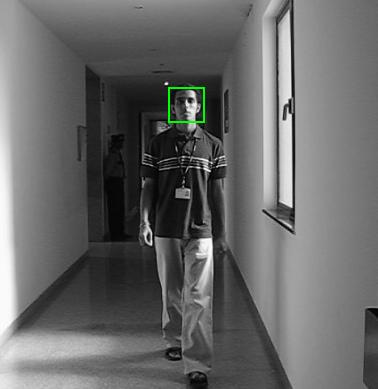}
&\includegraphics[height=20mm,width=24mm]{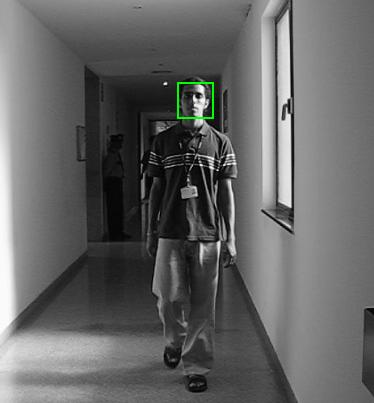}
&\includegraphics[height=20mm,width=24mm]{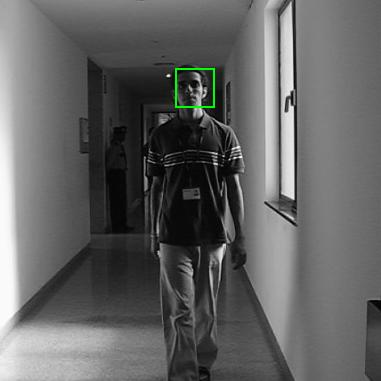}
&\includegraphics[height=20mm,width=24mm]{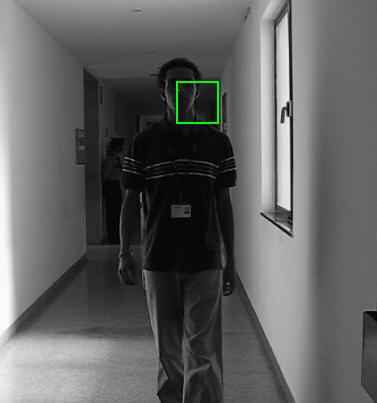}
\end{tabular}
}
\subfigure[tracking results using PaFiMoCS]{
\begin{tabular}{ccccc}
\includegraphics[height=20mm,width=24mm]{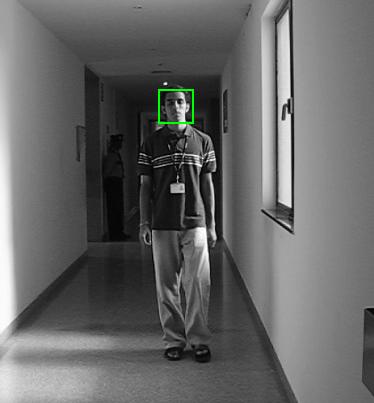}
&\includegraphics[height=20mm,width=24mm]{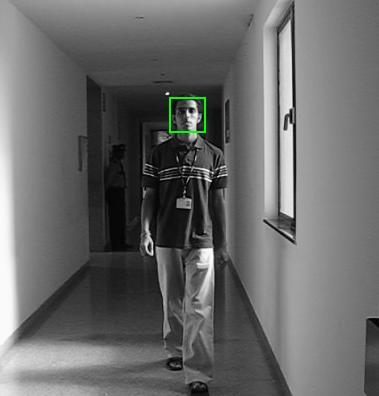}
&\includegraphics[height=20mm,width=24mm]{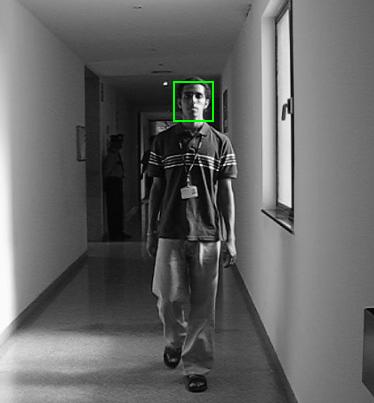}
&\includegraphics[height=20mm,width=24mm]{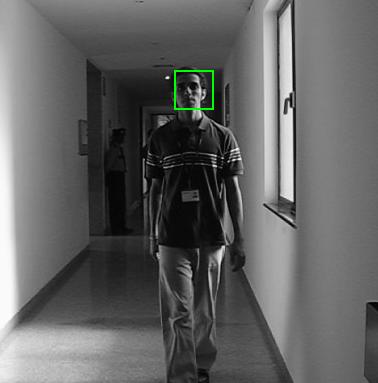}
&\includegraphics[height=20mm,width=24mm]{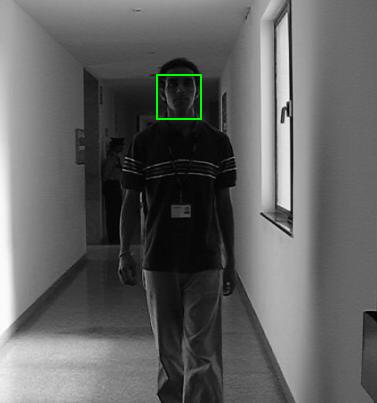}
\end{tabular}
}
\subfigure[tracking results using PF-MT-3]{
\begin{tabular}{ccccc}
\includegraphics[height=20mm,width=24mm]{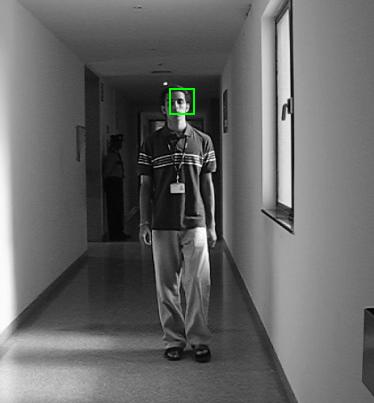}
&\includegraphics[height=20mm,width=24mm]{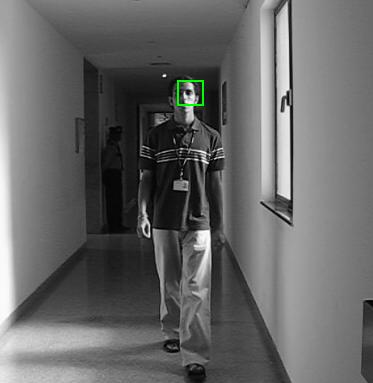}
&\includegraphics[height=20mm,width=24mm]{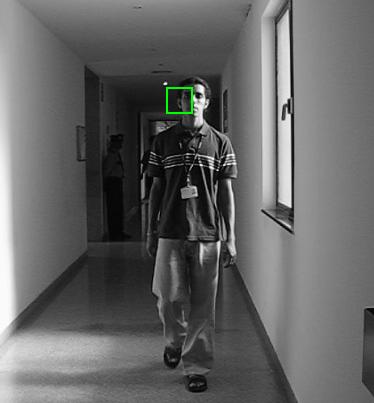}
&\includegraphics[height=20mm,width=24mm]{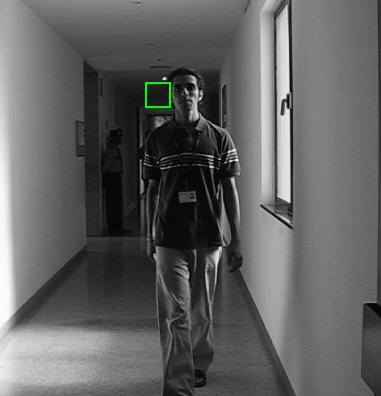}
&\includegraphics[height=20mm,width=24mm]{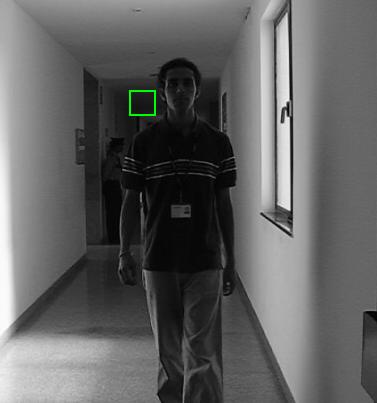}
\end{tabular}
}
\subfigure[tracking results using PF-MT-20]{
\begin{tabular}{ccccc}
\includegraphics[height=20mm,width=24mm]{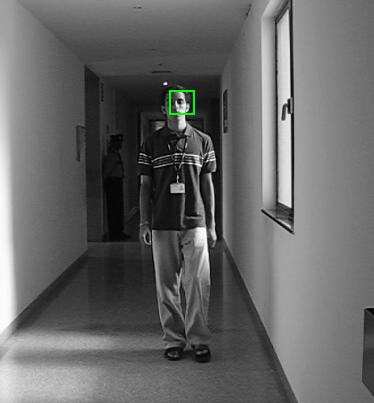}
&\includegraphics[height=20mm,width=24mm]{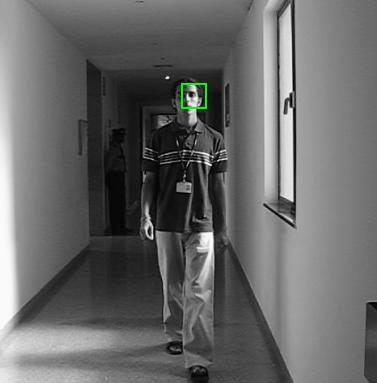}
&\includegraphics[height=20mm,width=24mm]{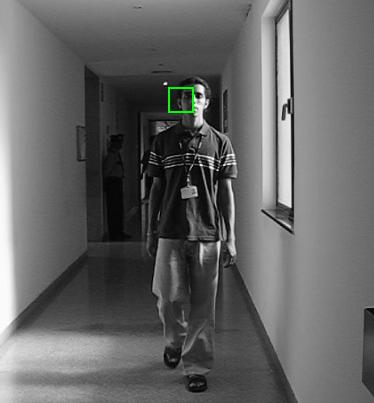}
&\includegraphics[height=20mm,width=24mm]{pfmt41_20.jpg}
&\includegraphics[height=20mm,width=24mm]{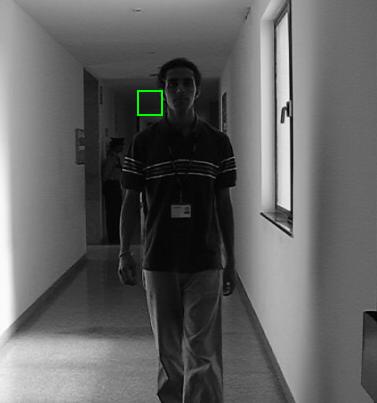}
\\ $t = 16$ & $t = 20$ & $t = 23$ & $t = 28$ & $t = 30$
\end{tabular}
}
\caption{{\small Tracking results for the indoor-walking sequence. 
We show the comparisons for frames 16, 20, 23, 28 and 30.
}}
\vspace{-0.1in}
\label{rahul_sam_vid}
\end{figure*}

From Fig \ref{blockplot}, we can see that the support set of all sequences does indeed contain many of the higher order Legendre polynomials. Polynomials up to the $16^{th}$ order ($k=32$) are present in all the three sequences. This explains why PF-MT run with only a 7-dimensional $\Lambda_t$ ($d=3$) fails to track these sequences [see Figs \ref{loc_error_video}, \ref{olivia_vid3}, \ref{rahul_sam_vid}]. This also indicates the need for using a Legendre dictionary with $d \ge 16$. To allow for occasional tracking errors, we used a dictionary with $d=20$ in our experiments.

We also repeated the above experiment with $\Phi$ being the Hadamard product of $I_0$ and the discrete Fourier transform matrix. The support size and support change size in the Fourier basis are plotted in Fig \ref{suppsize_Fourier}. As can be seen, for all the videos, the support size was much larger, between 70-90\% of $n_\lambda$. However, the support change size was smaller, between 10-20\% of the support size. Since the support size was so large, we did not use the Fourier basis in our experiments.

\Subsection{PaFiMoCS and PaFiMoCS-slow-support-change algorithms}
\label{pafimocs_illum}

With the state space model given above, the PaFiMoCS algorithm of Algorithm \ref{pafimocs_2_algo} and the PaFiMoCS-slow-support-change algorithm of Algorithm \ref{pafimocs_1_algo} for faster support changes apply directly. In either case, the cost function to minimize simplifies to
$$C(\Lambda) =  \frac{\|Y_t(\text{\text{ROI}}(U_t^{(i)}))-\text{vec}(I_0) - \Phi\Lambda\|_2^2}{ 2\sigma_{o}^2} + \beta \frac{ \|(\Lambda-\Lambda_{t-1}^{(i)})_{T}\|_2^2}{2\sigma_{l}^2} +  \gamma \| (\Lambda_{T^c})\|_1 $$
with $T= T_t^i$ in case of PaFiMoCS and $T=T_{t-1}^i$ in case of  PaFiMoCS-slow-support-change. Also, in the weighting step, $\text{OL}(U_t^i, \Phi\Lambda_t^i])$ is computed using (\ref{ol_gen_illum}).

Notice that for this problem, $g(Y_t,U_t, \Phi \Lambda_t)$ is indeed an affine function of $\Lambda_t$ and the noise $Z_t$ is Gaussian. As a result, the above cost function is convex in $\Lambda$ and thus easy to minimize using any of the standard convex solvers. We used CVX (CVX: Matlab software for disciplined convex programming, \url{http://cvxr.com/cvx}) in our implementations.

In order to deal with occlusions (outliers), one can use either of the approaches described earlier in Sec. \ref{outliers}. As explained there, using the second approach will retain the convexity of the cost function to minimize. With using that, the cost function to minimize becomes
$$C(\Lambda,O) =  \frac{\|Y_t(\text{\text{ROI}}(U_t^{(i)}))-\text{vec}(I_0) - \Phi\Lambda - O\|_2^2}{ 2\sigma_{o}^2} + \beta \frac{ \|(\Lambda-\Lambda_{t-1}^{(i)})_{T}\|_2^2}{2\sigma_{l}^2} +  \gamma \| \Lambda_{T^c}\|_1 + \gamma' \|O\|_1 $$

\Section{Experimental results}
\label{expts}

We show two types of experiments. In the first experiment, we took a face template and simulated a video sequence according to the models specified in Sec \ref{illum}. This allowed us to evaluate our algorithm and compare it against other PF methods for a dataset for which ground truth is available and for which one can generate multiple realizations of the sequence to compute the Monte Carlo based average performance. In the second experiment, we did comparisons on real video sequences containing significant spatial and temporal illumination changes. For one sequence in this experiment, we hand-marked the target's location in a 20-frame subsequence and treated this as ``ground truth" for quantitative performance evaluation. All comparisons also involve comparing visual displays of estimates of the target's bounding box.

We first discuss the results of the real video experiments in Sec \ref{real_expts}. The results of Monte Carlo evaluation using multiple simulated video sequences are described in Sec \ref{sim_expts}.

\Subsection{Results on Real Video Sequences}
\label{real_expts}




We show experiments on two different video sequences. In both cases, the goal was face tracking.
We compared our proposed algorithms with several other PF algorithms -- PF-MT  \cite{das_tip_11} (both using $d=3$ and $d=20$), Auxiliary-PF \cite{auxPF} (with $d=20$) and PF-Gordon \cite{gordon} (with $d=20$). All algorithms used 60 particles for all the videos.
As explained earlier and also in \cite{das_tip_11}, PF-Doucet \cite{doucet} cannot be implemented because the observation likelihood is not differentiable w.r.t. $U_t$ and hence the posterior mode cannot be computed. PF-MT with $d=3$ (PF-MT-3) is exactly the algorithm used in the experiments of \cite{das_tip_11}). Auxiliary-PF-3 and PF-Gordon-3 were already shown to fail for even simpler sequences in \cite{das_tip_11}, and in the simulation experiment that we show in Sec \ref{sim_expts}, and hence we do not repeat those comparisons here.

In the first sequence (outdoor-walking), the person walks under a tree and as the leaves of the tree move, different amounts of light fall on different parts of her face resulting in high frequency spatial variation of illumination that also changes with time. As a result, for many frames, many of the higher frequency (higher order) Legendre coefficients are also nonzero. We show the quantitative location error comparisons in Fig \ref{loc_error_video} and the bounding box display comparisons in Fig \ref{olivia_vid3}. The location error (LE) is computed as  $\text{LE} = \|(U_t)_{[1,2]} - (\hat{U}_t)_{[1,2]}\|_2$ where $U_t$ contains the translation of the centroid of the hand-marked bounding box and its scale w.r.t. the initial template and $\hat{U}_t$ is the tracked estimate of $U_t$ computed as the weighted mean of all the particles of $U_t$. To get the tracked bounding box shown in Fig \ref{olivia_vid3}, $(\hat{U}_t)_3$ is used to scale the initial template's bounding box and $(\hat{U}_t)_{[1,2]}$ is used to translate it. 

As can be seen from either Fig \ref{loc_error_video} or Fig \ref{olivia_vid3}, both PaFiMoCS and PaFiMoCS-slow-support-change are able to track the face in the entire sequence. For this sequence, most support changes are slow enough (see Fig \ref{suppchangeplot}) and hence both the PaFiMoCS algorithms had similar performance.
On the other hand, both PF-MT-20 and PF-MT-3 lose track, though the reasons are different. PF-MT-3 fails because it assumes that the illumination image can be accurately represented by only the first $7$ Legendre polynomials ($d=3$). However, as can be seen from Fig \ref{blockplot}, most frames of this sequence do contain significant energy in the higher order Legendre coefficients.
PF-MT-20 loses track because it does not exploit the sparsity or slow sparsity pattern change of $\Lambda_t$ and so, with probability one, it results in a dense solution for $\Lambda_t$, i.e. the energy gets distributed among all components of $\Lambda_t$. However, as can be seen from Fig \ref{suppsizeplot}, most of the energy of the true $\Lambda_t$ lies in only about 50\% of the 41 coefficients, while the others are zero or very small.
Aux-PF-20 and PF-Gordon-20 fail for two reasons. The first is the same as above, they also do not exploit sparsity or slow sparsity pattern change. Moreover, as explained in \cite{das_tip_11}, 60 particles is too few for all of these to be tracking on a 44 dimensional space. 

The second video sequence (indoor-walking) consisted of a person walking through a corridor across a window. The lighting is significantly more when he gets near the window and again lesser when he moves away from the window. Moreover, since the window is on the side of the person, the amount of light falling on different parts of his face is quite different (higher frequency spatial variations of illumination exist) and also changes with time. As can be seen from Fig \ref{rahul_sam_vid}, PF-MT-3 and PF-MT-20 lose track within a few frames for reasons similar to those explained above.
The illumination pattern changes are somewhat sudden when the person moves towards the window from a darker region of the corridor or vice versa, and so, as can be seen from Fig \ref{suppchangeplot}, the number of support changes at these times is significantly larger than in the outdoor-walking sequence. As a result PaFiMoCS-slow-support-change (Algorithm \ref{pafimocs_1_algo}) begins to lose track around $t=30$. However, PaFiMoCS (Algorithm \ref{pafimocs_2_algo}) remains in track always.

All algorithms used 60 particles for all the videos. The face template at the first frame was assumed known for all algorithms. For all the videos, the model parameters were set as\footnote{If training data is available, these parameters can be estimated by maximum likelihood estimation.} $\Sigma_u = \text{diag}(25,25,0.1)$,  $\sigma_l^2 = 1000$, $p_a = 0.06$, $\sigma_o^2 = 1000$. For PaFiMoCS, we used $\gamma= 0.7$ and $\beta=1$.  These values were selected after some experimentation using the approach suggested in \cite[Section V-B]{regmodbpdn}. As suggested in \cite{isitmodcs}, the support estimation threshold $\alpha$ was chosen as the 99\% energy threshold of the estimated illumination vector. For PaFiMoCS-slow-support-change, we used $\gamma= 0.5$ and $\beta=1$. A smaller value of $\gamma$ is needed for PaFiMoCS-slow-support-change because of the following reason. In this case we do not importance sample on the support. In the mode tracking step we minimize (\ref{pafimocs_cost_large_supp}) which uses the previous support particle $T_{t-1}^i$. Hence, a larger number of support additions may be needed in this case to get a support estimate that is close to the true $T_t$, i.e. $\Lambda_{(T_{t-1}^i)^c}$ may need to be less sparse.

\Subsection{Results on Simulated Video Sequences}
\label{sim_expts} 

A video sequence of a moving target with spatially varying illumination change was generated as follows. We let $I_0$ be a given face template. The vectors $\underline{i_0}, \underline{j_0}$ contained its x and y coordinates and $\bar{i}_0, \bar{j}_0$ contained the x,y coordinates of its centroid. Starting with $U_0 =  {\bf 0}$,  at time time $t$, the motion vector $U_t$ was generated according the random walk model given in (\ref{sysmod3}) with $\Sigma_u = \text{diag}(0.5,0.5,0)$. To keep things simple, we used a zero value for the scale variance, i.e. we simulated only x-y translation of the template.
The illumination vector $\Lambda_t$ was generated as follows. The initial support set $T_0$ contained five uniformly randomly selected indices from $[1,2,\dots (2d+1)]$ with $d=20$. Every five frames, the support set $T_t$ was changed according to (\ref{sysmod1}) with $p_a = 0.03$, $p_r = 0.216$. This ensured that the expected support size is 5 at each time and the expected support change size is 1 at each time. The illumination vector on the support $T_t$, $(\Lambda_t)_{T_t}$ was generated according to the random walk model of (\ref{sysmod2}) with $\sigma_l^2 = 0.01$ and initialized with $(\Lambda_0)_{T_0} = {\bf 0}$.
With $U_t$ generated as above, the ROI for the template at time $t$, $\text{ROI}(U_t)$, was computed using (\ref{roi}). The Legendre dictionary with $d=20$ was computed using (\ref{def_phi}). The observed image $Y_t$ was then generated using (\ref{yt_mod}) with $\sigma_o^2=1$. As given there, the pixels outside the ROI, i.e. those in $\text{ROI}(U_t)^c$, are assumed to be due to clutter and are modeled as being i.i.d. uniformly distributed between zero and 255.

\begin{figure*}[!t]
\centerline{
\subfigure[a simulated video sequence]{
\label{sim_seq_vid}
\begin{tabular}[c]{cccc}
\includegraphics[height=25mm,width=35mm]{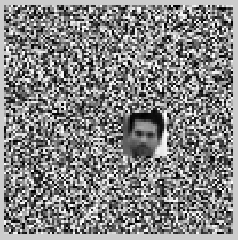}
&\includegraphics[height=25mm,width=35mm]{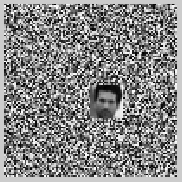}
&\includegraphics[height=25mm,width=35mm]{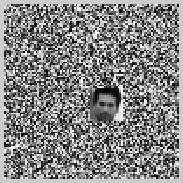}
&\includegraphics[height=25mm,width=35mm]{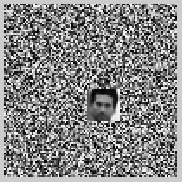}
\\
$t=6$ & $t=20$ & $t=35$ & $t=45$ 
\end{tabular}
}
}
\centerline{
\subfigure[NMSE of $(U_t, \Lambda_t)$]{  
\label{sim_seq_error}
\begin{tabular}[c]{c}
\includegraphics[height=45mm,width=80mm]{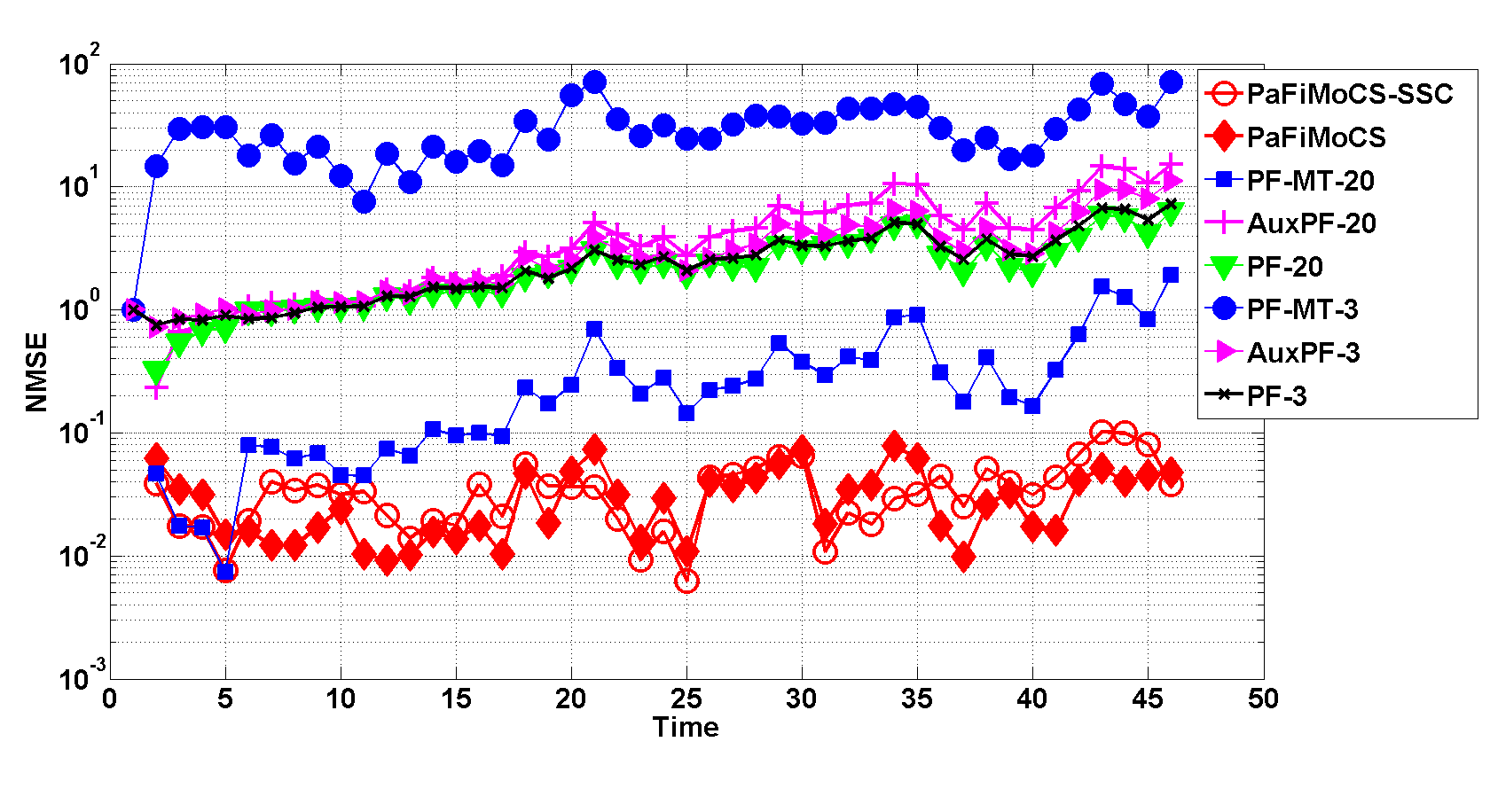}
\end{tabular}
}
}
\vspace{-0.15in}
\caption{{\small 
In the NMSE plot, PaFiMoCS-SSC refers to PaFiMoCS-slow-support-change.
}}
\vspace{-0.2in}
\label{sim_seq}
\end{figure*}

We generated 50 different video sequences using the above approach. An example sequence is shown in Fig \ref{sim_seq_vid}. Each sequence was tracked using PaFiMoCS and PaFiMoCS-slow-support-change with $d=20$ as well as using PF-MT  \cite{das_tip_11} (both using $d=3$ and $d=20$), Auxiliary-PF \cite{auxPF} (with $d=3$ and $d=20$) and PF-Gordon \cite{gordon} (with $d=3$ and $d=20$). All algorithms used 100 particles. In  Fig \ref{sim_seq_error}, we plot the normalized mean squared error (NMSE), $\text{NMSE}(t) := \frac{\E[\|U_t - \hat{U}_t\|_2^2 + \|\Lambda_t - \hat{\Lambda}_t\|_2^2]}{\E[\|U_t\|_2^2 + \|\Lambda_t\|_2^2]}$ against time. Here $\hat{U}_t$ and $\hat{\Lambda}_t$ are computed as the weighted means of all the particles of $U_t$ and $\Lambda_t$ respectively. The expectation is computed by averaging over the 50 Monte Carlo realizations of the sequence.

As can be seen, PaFiMoCS (Algorithm \ref{pafimocs_2_algo}) remains in track with stable and small error throughout. PaFiMoCS-slow-support-change (Algorithm \ref{pafimocs_1_algo}) also remains in track, but its errors are slightly larger because occasionally the number of support changes was large. PF-MT-3 \cite{das_tip_11} loses track because it assumes that only the first 7 Legendre polynomials are sufficient to represent the illumination image. However, we know from our simulation that the support of the illumination vector is equally likely to contain any element from $[1,2,\dots 41]$ (not just the first 7). On the other hand, as explained earlier, PF-MT-20 loses track because it assumes that $\Lambda_t$ is a dense vector, i.e. all of its 41 components are part of the support at all times. Aux-PF-20 and Aux-PF-3 as well as of PF-Gordon-20 and PF-Gordon-3 lose track due to similar reasons as above, and also because these need many more than 100 particles to even track on a 10 dimensional state space.

All algorithms used 100 particles. For PaFiMoCS, after some experimentation using the approach of \cite[Section V-B]{regmodbpdn}, we used $\gamma=0.7$ and $\beta=0.4$. As suggested in \cite{isitmodcs}, the support estimation threshold $\alpha$ was chosen as the 99\% energy threshold of the estimated illumination vector. 
For PaFiMoCS-slow-support-change, we used $\gamma= 0.5$, $\beta=0.4$, and $\alpha$ was again set as above. A smaller value of $\gamma$ is used for PaFiMoCS-slow-support-change for the same reason as that explained in Sec \ref{real_expts}. 

\Section{Conclusions and Future Work}
\label{conclude}
In this work, we studied the problem of tracking a time sequence of sparse spatial signals with changing sparsity patterns, e.g. illumination, as well as other unknown states, e.g. motion states, from a sequence of nonlinear observations corrupted by (possibly) non-Gaussian noise. A key application where this problem occurs is in tracking moving objects across spatially varying illumination change. In this case, the motion states form the small dimensional state vector, while the illumination ``image" (illumination at each pixel in the image) is the sparse spatial signal with slowly changing sparsity patterns.
%
We proposed a novel solution approach called Particle Filtered Modified-CS (PaFiMoCS). The key idea of PaFiMoCS is to importance sample for the small dimensional state vector, while replacing importance sampling by slow sparsity pattern change constrained posterior mode tracking for recovering the sparse spatial signal. We studied the illumination-motion tracking problem in detail and showed how to design PaFiMoCS for it.  Extensive experiments on both simulated as well as on real videos with significant illumination changes demonstrated the superiority of PaFiMoCS as compared with existing work.

Future work will involve designing PaFiMoCS using more sophisticated pose and appearance change models from recent work \cite{amitRC_2007,yuan_2008,kumar_2008}. 
A second goal of future work will be to study other visual tracking applications where the above problem occurs, e.g. tracking moving and deforming objects from cluttered and/or low contrast imagery.

\section*{Acknowledgement}
We would like thank Tyler Stapler for helping with the video collection required for this paper. 

\bibliographystyle{IEEEbib}
\bibliography{tipnewpfmt,shape_PAMI,U:/Proposals/Oct10/tipnewpfmt_kfcsfullpap,U:/Proposals/Oct10/tipnewpfmt_cq,pafimocs_new}

\begin{thebibliography}{10}

\bibitem{pafimocs_asilomar}
S.~Das and N.~Vaswani,
\newblock ``Particle filtered modified compressive sensing (pafimocs) for
  tracking signal sequences,''
\newblock in {\em Asilomar Conf. Signals, Systems and Computers}, 2010.

\bibitem{candes}
E.~Candes, J.~Romberg, and T.~Tao,
\newblock ``Robust uncertainty principles: Exact signal reconstruction from
  highly incomplete frequency information,''
\newblock {\em IEEE Trans. Info. Th.}, vol. 52(2), pp. 489--509, February 2006.

\bibitem{decodinglp}
E.~Candes and T.~Tao,
\newblock ``Decoding by linear programming,''
\newblock {\em IEEE Trans. Info. Th.}, vol. 51(12), pp. 4203 -- 4215, Dec.
  2005.

\bibitem{donoho}
D.~Donoho,
\newblock ``Compressed sensing,''
\newblock {\em IEEE Trans. on Information Theory}, vol. 52(4), pp. 1289--1306,
  April 2006.

\bibitem{kfcsicip}
N.~Vaswani,
\newblock ``Kalman filtered compressed sensing,''
\newblock in {\em IEEE Intl. Conf. Image Proc. (ICIP)}, 2008.

\bibitem{just_lscs}
N.~Vaswani,
\newblock ``{LS-CS-residual (LS-CS): Compressive Sensing on Least Squares
  residual},''
\newblock {\em IEEE Trans. Sig. Proc.}, vol. 58(8), pp. 4108--4120, August
  2010.

\bibitem{isitmodcs}
N.~Vaswani and W.~Lu,
\newblock ``Modified-cs: Modifying compressive sensing for problems with
  partially known support,''
\newblock {\em IEEE Trans. Sig. Proc.}, vol. 58(9), pp. 4595--4607, September
  2010.

\bibitem{hassibi}
A.~Khajehnejad, W.~Xu, A.~Avestimehr, and B.~Hassibi,
\newblock ``Weighted $\ell_{1}$ minimization for sparse recovery with prior
  information,''
\newblock {\em IEEE Trans. Sig. Proc.}, 2011.

\bibitem{modblockcs_stojnic}
M.~Stojnic,
\newblock ``Block-length dependent thresholds for $\ell_2 /
  \ell_1$-optimization in block-sparse compressed sensing,''
\newblock in {\em ICASSP}, 2010.

\bibitem{regmodbpdn}
W.~Lu and N.~Vaswani,
\newblock ``Regularized modified bpdn for noisy sparse reconstruction with
  partial erroneous support and signal value knowledge,''
\newblock {\em IEEE Trans. Sig. Proc.}, January 2012.

\bibitem{ibm}
A.~Carmi, P.~Gurfil, and D.~Kanevsky,
\newblock ``Methods for sparse signal recovery using kalman filtering with
  embedded pseudo-measurement norms and quasi-norms,''
\newblock {\em IEEE Trans. Sig. Proc.}, April 2010.

\bibitem{sparse_dynamic_sys}
D.~Sejdinovic, C.~Andrieu, and R.~Piechocki,
\newblock ``Bayesian sequential compressed sensing in sparse dynamical
  systems,''
\newblock in {\em Allerton Conf. Communication, Control, and Computing}, 2010.

\bibitem{schniter_track}
J.~Ziniel, L.~C. Potter, and P.~Schniter,
\newblock ``Tracking and smoothing of time-varying sparse signals via
  approximate belief propagation,''
\newblock in {\em Asilomar Conf. on Sig. Sys. Comp.}, 2010.

\bibitem{ashwin_eccv_2010}
A.~C. Sankaranarayanan, P.~K. Turaga, R.~G. Baraniuk, and R.~Chellappa,
\newblock ``Compressive acquisition of dynamic scenes,''
\newblock in {\em Eur. Conf. on Comp. Vis. (ECCV)}, 2010.

\bibitem{zhang_rao}
Z.~Zhang and B.~D. Rao,
\newblock ``Sparse signal recovery with temporally correlated source vectors
  using sparse bayesian learning,''
\newblock {\em IEEE J. Sel. Topics Sig. Proc., Special Issue on Adaptive Sparse
  Representation of Data and Applications in Signal and Image Processing}, vol.
  5, no. 5, pp. 912--926, Sept 2011.

\bibitem{romberg_ciss11}
A.~Charles, M.~S. Asif, J.~Romberg, and C.~Rozell,
\newblock ``Sparsity penalties in dynamical system estimation,''
\newblock in {\em Conf. Info. Sciences and Systems}, 2011.

\bibitem{unscented_cs}
A.~Y. Carmi and L.~Mihailova,
\newblock ``Unscented compressed sensing,''
\newblock in {\em IEEE Intl. Conf. Acoustics, Speech, Sig. Proc. (ICASSP)},
  2012.

\bibitem{doucet}
A.~Doucet,
\newblock ``On sequential monte carlo sampling methods for bayesian
  filtering,''
\newblock in {\em Technical Report CUED/F-INFENG/TR. 310, Cambridge University
  Department of Engineering}, 1998.

\bibitem{gordon}
N.~J. Gordon, D.~J. Salmond, and A.~F.~M. Smith,
\newblock ``Novel approach to nonlinear/nongaussian bayesian state
  estimation,''
\newblock {\em IEE Proceedings-F (Radar and Signal Processing)}, pp.
  140(2):107--113, 1993.

\bibitem{das_tip_11}
S.~Das, A.~Kale, and N.~Vaswani,
\newblock ``Particle filter with mode tracker (pf-mt) for visual tracking
  across illumination changes,''
\newblock {\em IEEE Trans. Image Proc.}, April 2012.

\bibitem{gpf}
J.~H. Kotecha and P.~M. Djuric,
\newblock ``Gaussian particle filtering,''
\newblock {\em IEEE Trans. Sig. Proc.}, pp. 2592--2601, Oct 2003.

\bibitem{gspf}
J.~H. Kotecha and P.~M. Djuric,
\newblock ``Gaussian sum particle filtering,''
\newblock {\em IEEE Trans. Sig. Proc.}, pp. 2602--2612, Oct 2003.

\bibitem{mcip}
A.~Doucet, N.~deFreitas, and N.~Gordon, Eds.,
\newblock {\em Sequential Monte Carlo Methods in Practice},
\newblock Springer, 2001.

\bibitem{gust_nord}
T.~Schn, F.~Gustafsson, and P.~Nordlund,
\newblock ``Marginalized particle filters for nonlinear state-space models,''
\newblock {\em IEEE Trans. Sig. Proc.}, 2005.

\bibitem{chen_liu}
R.~Chen and J.S. Liu,
\newblock ``Mixture kalman filters,''
\newblock {\em Journal of the Royal Statistical Society}, vol. 62(3), pp.
  493--508, 2000.

\bibitem{pfmtpap}
N.~Vaswani,
\newblock ``Particle filtering for large dimensional state spaces with
  multimodal observation likelihoods,''
\newblock {\em IEEE Trans. Sig. Proc.}, pp. 4583--4597, October 2008.

\bibitem{chang_TIP2008}
W.~Y. Chang, C.~S. Chen, and Y.~D. Jian,
\newblock ``Visual tracking in high-dimensional state space by
  appearance-guided particle filtering,''
\newblock {\em Trans. Img. Proc.}, vol. 17, no. 7, pp. 1154--1167, July 2008.

\bibitem{mihaylova_ICASSP2011}
L.~Mihaylova and A.~Carmi,
\newblock ``Particle algorithms for filtering in high dimensional state spaces:
  A case study in group object tracking,''
\newblock in {\em ICASSP}, 2011, pp. 5932--5935.

\bibitem{mihaylova_TITS2012}
L.~Mihaylova, A.~Hegyi, A.~Gning, and R.~K. Boel,
\newblock ``Parallelized particle and gaussian sum particle filters for
  large-scale freeway traffic systems,''
\newblock {\em IEEE Transactions on Intelligent Transportation Systems}, vol.
  13, no. 1, pp. 36--48, 2012.

\bibitem{partas_2010}
E.~Besada-Portas, S.~M. Plis, J.~M. de~la Cruz, and T.~Lane,
\newblock ``Adaptive parallel/serial sampling mechanisms for particle filtering
  in dynamic bayesian networks,''
\newblock in {\em Proceedings of the 2010 European conference on Machine
  learning and knowledge discovery in databases: Part I}, Berlin, Heidelberg,
  2010, ECML PKDD'10, pp. 119--134, Springer-Verlag.

\bibitem{Djuric_ICASSP2011}
B.~Balasingam, M.~Bolic, P.~M. Djuric, and J.~M\'{\i}guez,
\newblock ``Efficient distributed resampling for particle filters,''
\newblock in {\em ICASSP}, 2011, pp. 3772--3775.

\bibitem{zhang_ICASSP2012}
N.~Zheng, Y.~Pan, and X.~Yan,
\newblock ``Hierarchical resampling architecture for distributed particle
  filters,''
\newblock in {\em International Conference on Acoustics, Speech, and Signal
  Processing}, 2012, ICASSP'12.

\bibitem{sparse_kernel_TIP2010}
A.~Banerjee and P.~Burlina,
\newblock ``Efficient particle filtering via sparse kernel density
  estimation,''
\newblock {\em IEEE Transactions on Image Processing}, vol. 19, no. 9, pp.
  2480--2490, 2010.

\bibitem{ahuja_CVPR2012}
Z.~Tianzhu, B.~Ghanem, and N.~Ahuja,
\newblock ``Robust visual tracking via multi-task sparse learning,''
\newblock in {\em Proceedings of the 2012 IEEE Conference on Computer Vision
  and Pattern Recognition}, 2012, CVPR '12.

\bibitem{compressive_tracking_ECCV2012}
K.~Zhang, L.~Zhang, and M.~H. Yang,
\newblock ``Real-time compressive tracking,''
\newblock in {\em Proceedings of the 2012 European Conference on Computer
  Vision}, 2012, ECCV '12.

\bibitem{VisualTrackingl1}
X.~Mei and H.~Ling,
\newblock ``Robust visual tracking using $\ell_1$ minimization,''
\newblock in {\em ICCV}, 2009.

\bibitem{VisualTrackingCS}
H.~Li, C.~Shen, and Q.~Shi,
\newblock ``Real-time visual tracking using compressive sensing,''
\newblock in {\em CVPR}, 2011.

\bibitem{zhuang_eccv10}
B.~Liu, L.~Yang, J.~Huang, P.~Meer, L.~Gong, and C.Kulikowski,
\newblock ``Robust and fast visual tracking with two stage sparse
  optimization,''
\newblock in {\em ECCV}, 2010.

\bibitem{weiss_iccv}
Y.~Weiss,
\newblock ``Deriving intrinsic images from image sequences,''
\newblock in {\em IEEE Intl. Conf. on Computer Vision (ICCV)}, 2001.

\bibitem{kalecvpr}
A.~Kale and C.~Jaynes,
\newblock ``A joint illumination and shape model for visual tracking,''
\newblock in {\em IEEE Conf. on Comp. Vis. Pat. Rec. (CVPR)}, 2006, pp.
  602--609.

\bibitem{amitRC_2007}
Yilei Xu and A.~K. Roy-Chowdhury,
\newblock ``Integrating motion, illumination, and structure in video sequences
  with applications in illumination-invariant tracking,''
\newblock {\em IEEE Trans. Pattern Anal. Machine Intell.}, 2007.

\bibitem{ross_2008}
David~A. Ross, Jongwoo Lim, Ruei-Sung, and Lin Ming-Hsuan Yang,
\newblock ``Incremental learning for robust visual tracking,''
\newblock {\em Intl. Journal Comp. Vis.}, vol. 77, pp. 125--141, 2008.

\bibitem{soatto_2008}
J.D. Jackson, A.J. Yezzi, and S.~Soatto,
\newblock ``Dynamic shape and appearance modeling via moving and deforming
  layers,''
\newblock {\em Intl. Journal Comp. Vis.}, vol. 79, no. 1, pp. 71--84, August
  2008.

\bibitem{yuan_2008}
Yuan Li, Haizhou Ai, Takayoshi Yamashita, Shihong Lao, and Masato Kawade,
\newblock ``Tracking in low frame rate video: A cascade particle filter with
  discriminative observers of different life spans,''
\newblock {\em IEEE Transactions on Pattern Analysis and Machine Intelligence},
  vol. 30, no. 10, pp. 1728--1740, 2008.

\bibitem{kumar_2008}
Minyoung Kim, S~Kumar, V.~Pavlovic, and H~Rowley,
\newblock ``Face tracking and recognition with visual constraints in real-world
  videos,''
\newblock in {\em IEEE Conf. on Comp. Vis. Pat. Rec. (CVPR)}, June 2008.

\bibitem{jacobs_jacobs}
Jacobs~R. Basri and D.~Jacobs,
\newblock ``Lambertian reflectance and linear subspaces,''
\newblock {\em IEEE Trans. Pattern Anal. Machine Intell.}, vol. 25, no. 2, pp.
  218--233, 2003.

\bibitem{belmur_cone}
Belkrieg~P. Belhumeur and D.~J. Kriegman,
\newblock ``What is the set of images of an object under all possible
  illumination conditions,''
\newblock vol. 28, no. 3, pp. 1--16, 1998.

\bibitem{hager_bell}
G.~Hager and P.~Belhumeur,
\newblock ``Efficient region tracking with parametric models of geometry and
  illumination,''
\newblock {\em IEEE Trans. Pattern Anal. Machine Intell.}, vol. 20, no. 10, pp.
  1025–1039, 1998.

\bibitem{ramamoorthi_pca}
R.~Ramamoorthi,
\newblock ``Analytic pca construction for theoretical analysis of lighting
  variability in images of lambertian object,''
\newblock {\em IEEE Trans. Pattern Anal. Machine Intell.}, vol. 24, no. 10, pp.
  1--12, 2002.

\bibitem{condensation}
M.~Isard and A.~Blake,
\newblock ``{Condensation: Conditional Density Propagation for Visual
  Tracking},''
\newblock {\em Intl. Journal Comp. Vis.}, pp. 5--28, 1998.

\bibitem{error_correction_PCP_l1}
John Wright and Yi~Ma,
\newblock ``Dense error correction via l1-minimization,''
\newblock {\em IEEE Transactions on Information Theory}, 2009.

\bibitem{auxPF}
M.~Pitt and N.~Shephard,
\newblock ``Filtering via simulation: auxiliary particle filters,''
\newblock {\em J. Amer. Stat. Assoc}, vol. 94, pp. 590–599, 1999.

\bibitem{contour}
N.~Vaswani, Y.~Rathi, A.~Yezzi, and A.~Tannenbaum,
\newblock ``Deform pf-mt : Particle filter with mode tracker for tracking
  non-affine contour deformations,''
\newblock {\em IEEE Trans. Image Proc.}, April 2010.

\end{thebibliography}

\end{document}